\documentclass[10pt,twocolumn,letterpaper]{article}

\usepackage{iccv}
\usepackage{titling}
\usepackage{times}
\usepackage{epsfig}
\usepackage{graphicx}
\usepackage{amsmath}
\usepackage{amssymb}
\usepackage{graphicx}
\usepackage{amsmath}
\usepackage{amssymb}
\usepackage{booktabs}
\usepackage{multirow}
\usepackage{bm}
\usepackage{wrapfig}
\usepackage[ruled]{algorithm2e}
\usepackage{amsthm,amsmath,amssymb}
\usepackage{mathrsfs}
\usepackage{colortbl}  
\usepackage{xcolor}
\usepackage{array}   
\usepackage{arydshln}

\usepackage[pagebackref=true,breaklinks=true,letterpaper=true,colorlinks,bookmarks=false]{hyperref}

\iccvfinalcopy 

\def\iccvPaperID{1111} 

\begin{document}

{\title{\textbf{Monte Carlo Linear Clustering with Single-Point Supervision is Enough for Infrared Small Target Detection}}

\author{Boyang Li$^{1}$, Yingqian Wang$^{1}$, Longguang Wang$^{2}$, Fei Zhang$^{3}$, \\Ting Liu$^{1}$, Zaiping Lin$^{1}$, Wei An$^{1}$, Yulan Guo$^{1}$\\
$^{1}$National University of Defense Technology, $^{2}$Aviation University of Air Force, \\$^{3}$Shanghai Jiao Tong University\\
}
\date{}
\maketitle

\begin{abstract}
Single-frame infrared small target (SIRST) detection aims at separating small targets from clutter backgrounds on infrared images. Recently, deep learning based methods have achieved promising performance on SIRST detection, but at the cost of a large amount of training data with expensive pixel-level annotations. To reduce the annotation burden,  we propose the first method to achieve SIRST detection with single-point supervision. The core idea of this work is to recover the per-pixel mask of each target from the given single point label by using clustering approaches, which looks simple but is indeed challenging since targets are always insalient and accompanied with background clutters. To handle this issue, we introduce randomness to the clustering process by adding noise to the input images, and then obtain much more reliable pseudo masks by averaging the clustered results. Thanks to this ``Monte Carlo'' clustering approach, our method can accurately recover pseudo masks and thus turn arbitrary fully supervised SIRST detection networks into weakly supervised ones with only single point annotation. Experiments on four datasets demonstrate that our method can be applied to existing SIRST detection networks to achieve comparable performance with their fully-supervised counterparts, which reveals that single-point supervision is strong enough for SIRST detection. Our code will be available at: \url{https://github.com/YeRen123455/SIRST-Single-Point-Supervision}.

\end{abstract}


\section{Introduction}
\label{sec:intro}

Single-frame infrared small target (SIRST) detection has been widely used in many applications such as marine resource utilization \cite{DNANet, Shape-matter}, high-precision navigation \cite{3-anti-miss, 2-early-warning}, and ecological environment monitoring \cite{MDvsFA}. However, existing SIRST detection methods mainly rely on segmentation pipeline with pixel-level supervision. Pixel-level annotating is time-consuming and thus hinders the quick deployment of those large scale data-dependent scenarios. Therefore, it is in urgent need to alleviate the annotation burden while maintaining state-of-the-art performance in SIRST detection.

Point-level supervision, as an annotation-friendly form of supervision, can provide both category and localization information for each target. Previous works in general object segmentation utilized rich semantic information to design various training loss \cite{First-Single} and network modules \cite{Four-point1,Four-point2}, and continuously expand the single-point annotation to a pixel-level pseudo mask. Although having achieved promising performance, these methods heavily rely on prior semantic information (e.g., objectness prior in \cite{First-Single,CAM}, shape prior in \cite{Four-point1, Four-point2, Cell-point}), and generally need more than one point as supervision. However, infrared small targets generally occupy no more than 0.15\% area of the whole image \cite{SPIE} and lack texture, shape and color information. Consequently, previous semantic information-dependent methods cannot well adopt to this task.

\begin{figure}
\centering
\includegraphics[width=0.95\linewidth]{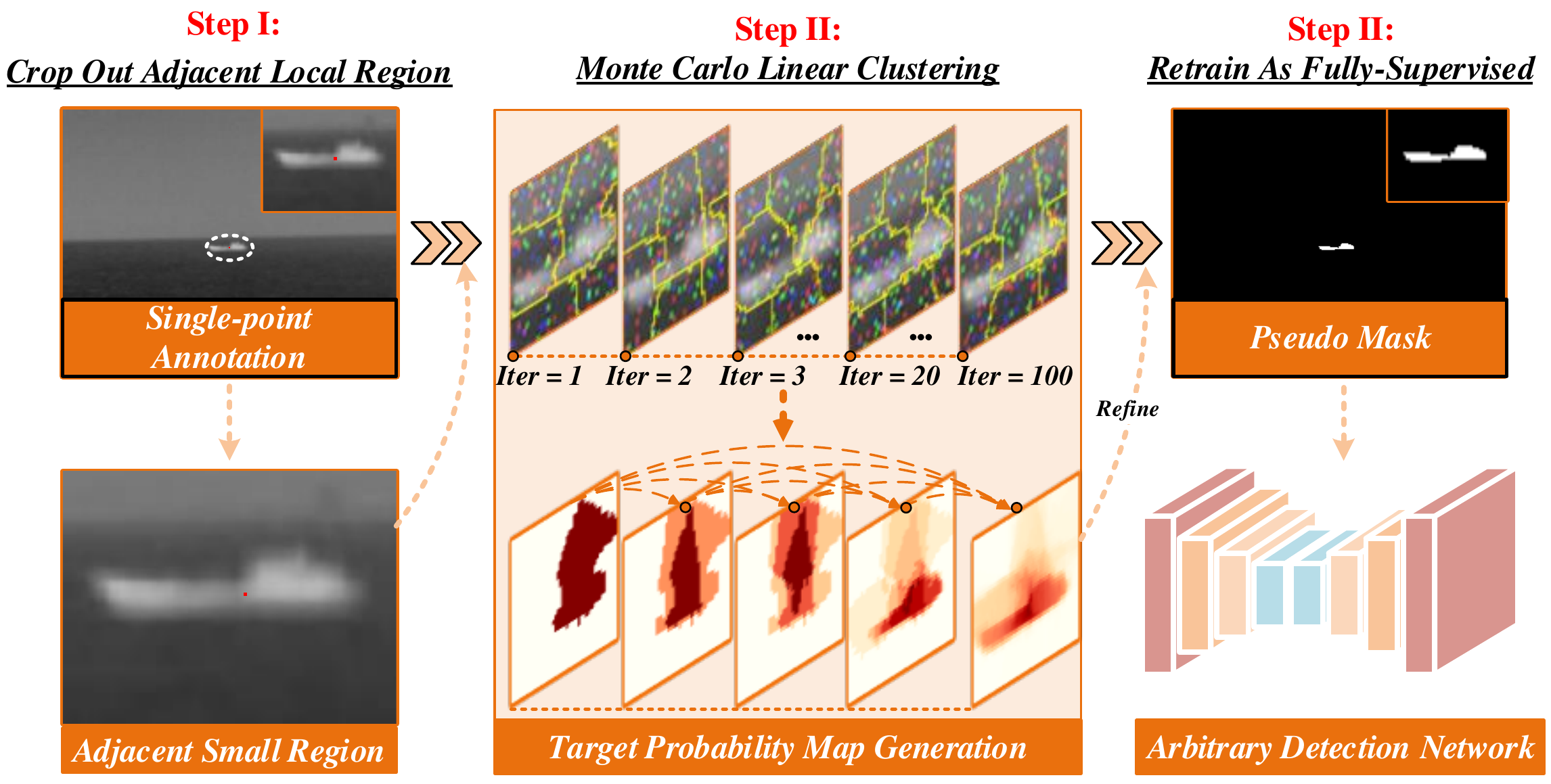}
\caption{Pipeline of our method. The proposed Monte Carlo linear clustering along with single-point annotation can produce high-quality pseudo mask (i.e., target probability map (TPM)).}\label{Fig_1}
\end{figure}

We notice that although infrared small targets lack semantic information from the perspective of whole image, they are quite salient in the local small region. Specifically, as shown in Fig.~\ref{Fig_2}, small targets in infrared images generally have high energy concentricity and exhibit significant gradient difference in their local regions. Moreover, limited semantic information (e.g., weak texture, shape, and color distribution) makes most point-like and spot-like targets exhibit similar energy distribution (e.g., Gaussian distribution). Consequently, it is straightforward to use a simple linear clustering approach (LCA) to separate the salient regions by measuring their color and spatial distances with adjacent backgrounds.

\begin{figure}
\centering
\includegraphics[width=0.90\linewidth]{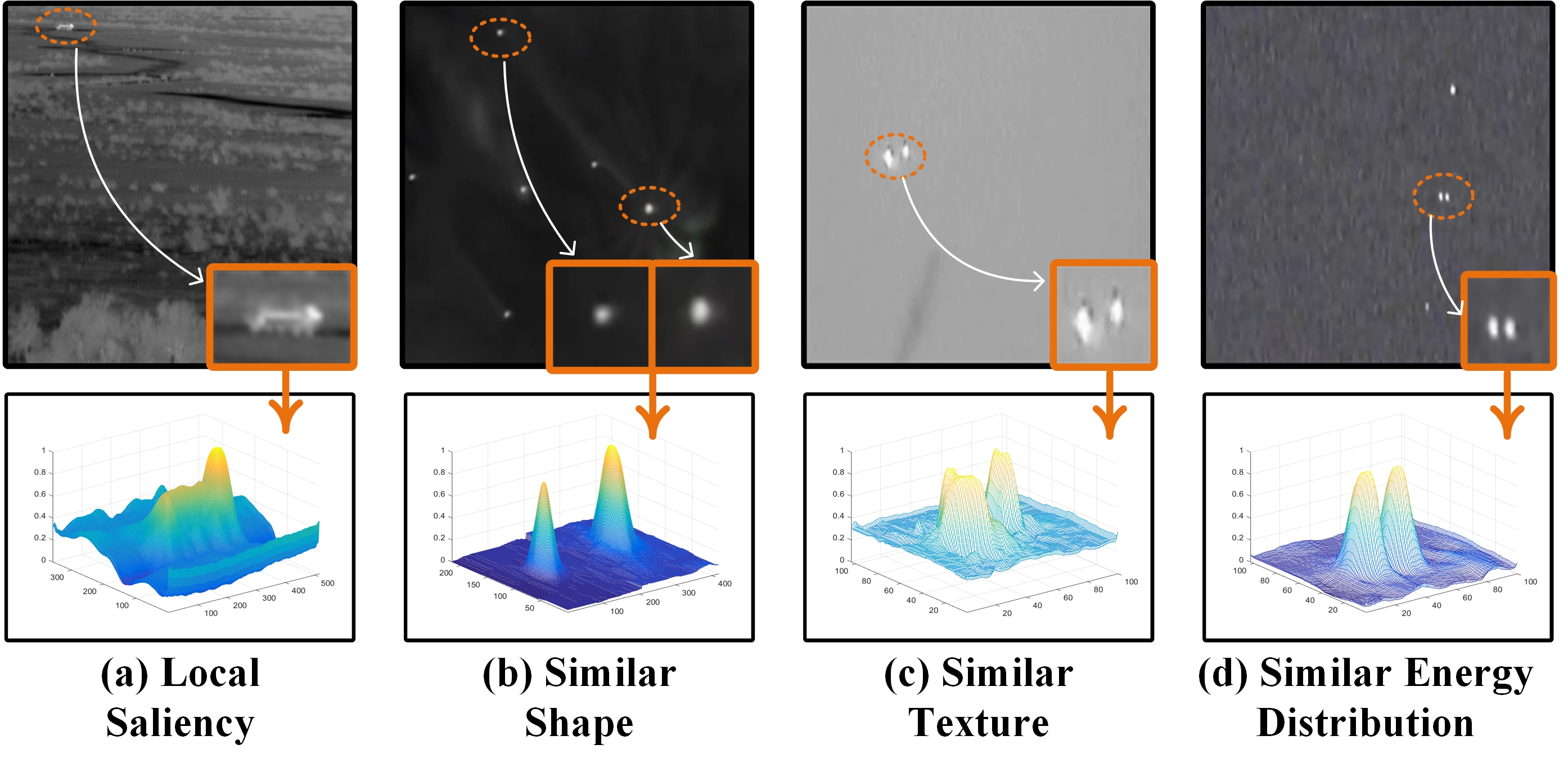}
\caption{Some inherent characteristics of SIRST observed from the small local regions (i.e., salient in local region, similar shape, texture, and energy distribution). }\label{Fig_2}
\end{figure}

However, LCA easily falls into local optimum and produces inaccurate segmentation results due to their fixed hyper-parameters. To handle this problem, we propose a Monte Carlo linear clustering (MCLC) method to regularize the clustering process and recover the reliable clustering result by repetitive random experiments. Specifically, we introduce randomness to the clustering process by adding noise to the input images. With the help of random noise, the distance between unexpected background region and target region can be significantly enlarged. The misclustered target regions can be pushed away from the false clustering center and thus return back the true clustering center. In this way, the single-point annotation can be gradually recovered as a much more reliable pixel-level target probability map (TPM) by averaging the clustered results. Pixels with larger values in TPM represent a higher probability of belonging to the target. Finally, we use the refined TPM to turn arbitrary fully supervised SIRST detection networks into weakly supervised ones with only single point annotation. Fig.~\ref{Fig_1} shows the overall pipeline of our method, and the main contributions are summarized as follows:

\begin{itemize}

\item To the best of our knowledge, this is the first single-point supervised method to achieve SIRST detection. A Monte Carlo linear clustering-based pipeline is proposed to achieve comparable performance with the fully supervised counterpart by using single-point annotaion.
\item  Inspired by the inherent characteristic of SIRST, a simple yet effective linear clustering approach with random noise-guided Monte Carlo regularization is proposed to coarsely extract and further refine the candidate target region.
\item Experiments on four public SIRST datasets demonstrate the effectiveness of our method. Ablation study reveals that pixel-level labels are not necessary for SIRST detection while single-point supervision is strong enough.
\end{itemize}

\section{Related Work}
\label{sec:formatting}

\subsection{Infrared Small Target Detection}

SIRST detection has been extensively investigated for decades. Early traditional paradigm achieves SIRST detection by measuring the discontinuity between targets and backgrounds. Typical methods include filtering-based methods \cite{4-tophat,5-maxmedian}, local contrast measure-based methods \cite{6-LCM,7-Robust-LCM,8-TLLCM,9-WSLLCM,Local_contrast_01,Local_contrast_02}, and low rank-based methods \cite{10-IPI,11-NRAM,12-RIPT,13-PSTNN,low_rank_01,low_rank_02}. Since real scenes are much more complex with dramatic changes in target size, shape, and clutter background, it is difficult to use handcrafted features and fixed hyper-parameters to handle such variations. To address this problem, recent deep learning-based methods learn trainable features in a data-driven manner and thus achieve better performance than traditional ones.

Existing deep learning-based methods can be divided into detection based methods and segmentation based methods. Since the pixel-level classification result is essential for the subsequent recognition task in SIRST, segmentation-based methods have attracted increasing attention recently. Dai et al. \cite{ACM} proposed the first segmentation-based network (i.e., ACM). They designed an asymmetric contextual module to aggregate features from shallow layers and deep layers. Then, Dai et al. \cite{ALCNet} improved ACM by introducing a dilated local contrast measure. Specifically, a feature cyclic shift scheme was designed to achieve a trainable local contrast measure.
After that, Zhang et al. \cite{Shape-matter} modeled SIRST detection as a shape detection task.  A taylor finite difference (TFD)-inspired edge block and a two-orientation attention aggregation (TOAA) block were proposed to capture precise shape of infrared targets. Recentlly, Li et al. \cite{DNANet} proposed a dense nested attention network (DNANet). A specifically-designed dense nested interactive module (DNIM) was proposed to both extract high-level information and maintain the response of small targets in deep layers.

Although the performance has been continuously improved by recent networks, existing deep learning-based methods rely on the fully-supervised training scheme with pixel-level annotations. Expensive labor cost makes existing methods hard to be deployed to large scale data-dependent tasks.

\subsection{Point-based Segmentation}

\begin{figure}
\centering
\includegraphics[width=0.90\linewidth]{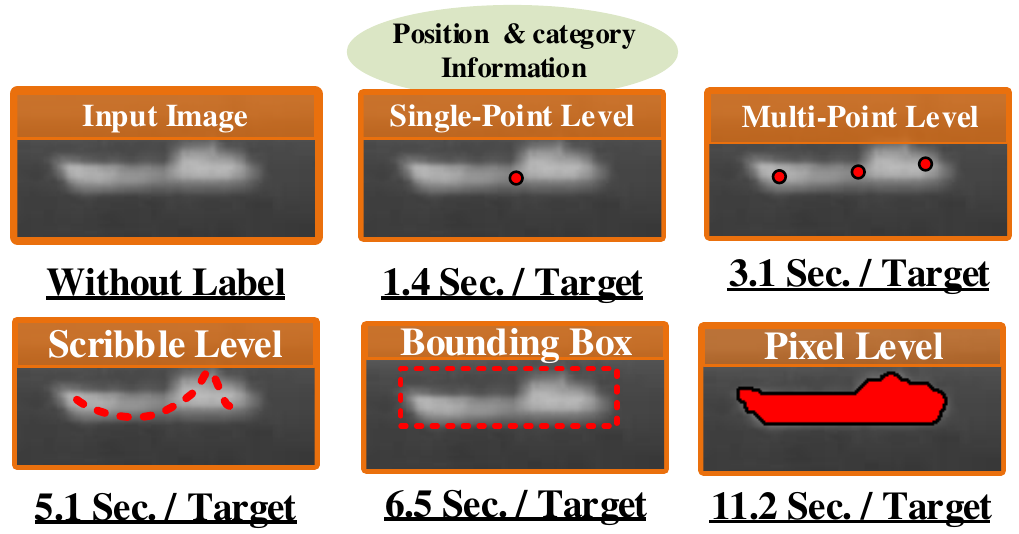}
\caption{Sample of input images, four common weakly supervised annotations (i.e., single-point, multi-point, scribble, bounding box), and one fully supervised pixel-level annotation.}\label{Fig_4}
\end{figure}

Bearman et al. \cite{First-Single} proposed the first single-point supervised semantic segmentation method. They incorporated the single-point supervision along with an objectness prior as the loss function to infer the extent of the object. Then, Papadopoulos et al. \cite{Four-point1} and Maninis et al. \cite{Four-point2} utilized four extreme points  (i.e., left-most, right-most, top, and bottom pixels) as supervision to further improve the quality of the pseudo mask. After that, Austin et al. \cite{PCAMS} followed the class activation maps (CAM) \cite{CAM} pipeline and used the point annotation to refine the quality of  pseudo labels. More recently, Li et al. \cite{20Point} utilized the semantical consistency property of general objects with 20 randomly annotated points to achieve comparable segmentation results with fully-supervised counterpart. Moreover, in the field of cell segmentation, Zhao et al. \cite{Cell-point} proposed the first single-point supervised segmentation method, in which two semantic prior-based training losses (i.e., divergence loss and consistency loss) were proposed to achieve high-quality cell segmentation.

Although promising performance has been achieved, existing works rely on rich semantic information (e.g., objectness prior in \cite{First-Single,CAM}, shape prior in \cite{Four-point1, Four-point2, Cell-point}) and most of them need multi-point annotation. Infrared small targets generally occupy no more than 0.15\% area of the whole image \cite{SPIE} and lack texture, shape and color information. Previous semantic information-dependent methods cannot be directly used for SIRST detection.

\section{Methodology}\label{methodology}

In this section, we first introduce the motivation. Then, a detailed illustration of the proposed linear clustering approach with Monte Carlo regularization is provided. Overall architecture of proposed method is shown in Fig.~\ref{Fig_3}.

\subsection{Motivation}\label{motivation}

We investigate the annotation cost of four common weakly-supervised annotations (i.e., single point \cite{Point}, multiple points\cite{Point, Four-point1, Four-point2}, scribbles \cite{Scribblesup}, bounding boxes\cite{simple-as}) by re-labelling the NUDT-SIRST \cite{DNANet} and NUAA-SIRST \cite{ACM} datasets. Fig. \ref{Fig_4} shows the average annotation time of four weakly-supervised and one fully-supervised approaches. Note that, single-point supervision can reduce about 87\% annotation time as compared to the pixel-level annotation approach, which motivates us to apply single-point annotation to SIRST detection. Please refer to the supplementary material  for more detailed analysis.

However, existing point-level supervised methods \cite{Point, Four-point1, Four-point2} are designed for general objects with rich semantic information (e.g., shape, texture, and color). As shown in Fig.~\ref{Fig_2}, the  characteristics of infrared small targets make the aforementioned semantic-based pipeline unsuitable for SIRST detection. Considering that infrared small targets are quite salient in their local regions and share similar shape, texture, and energy distribution. Intuitively, we can use simple linear clustering approaches (e.g., Super-pixel \cite{SLIC}, KMeans \cite{kmeans}) to separate the salient regions by comparing their color and spatial distances among these clusters.

\begin{figure*}
\centering
\includegraphics[width=0.85\linewidth]{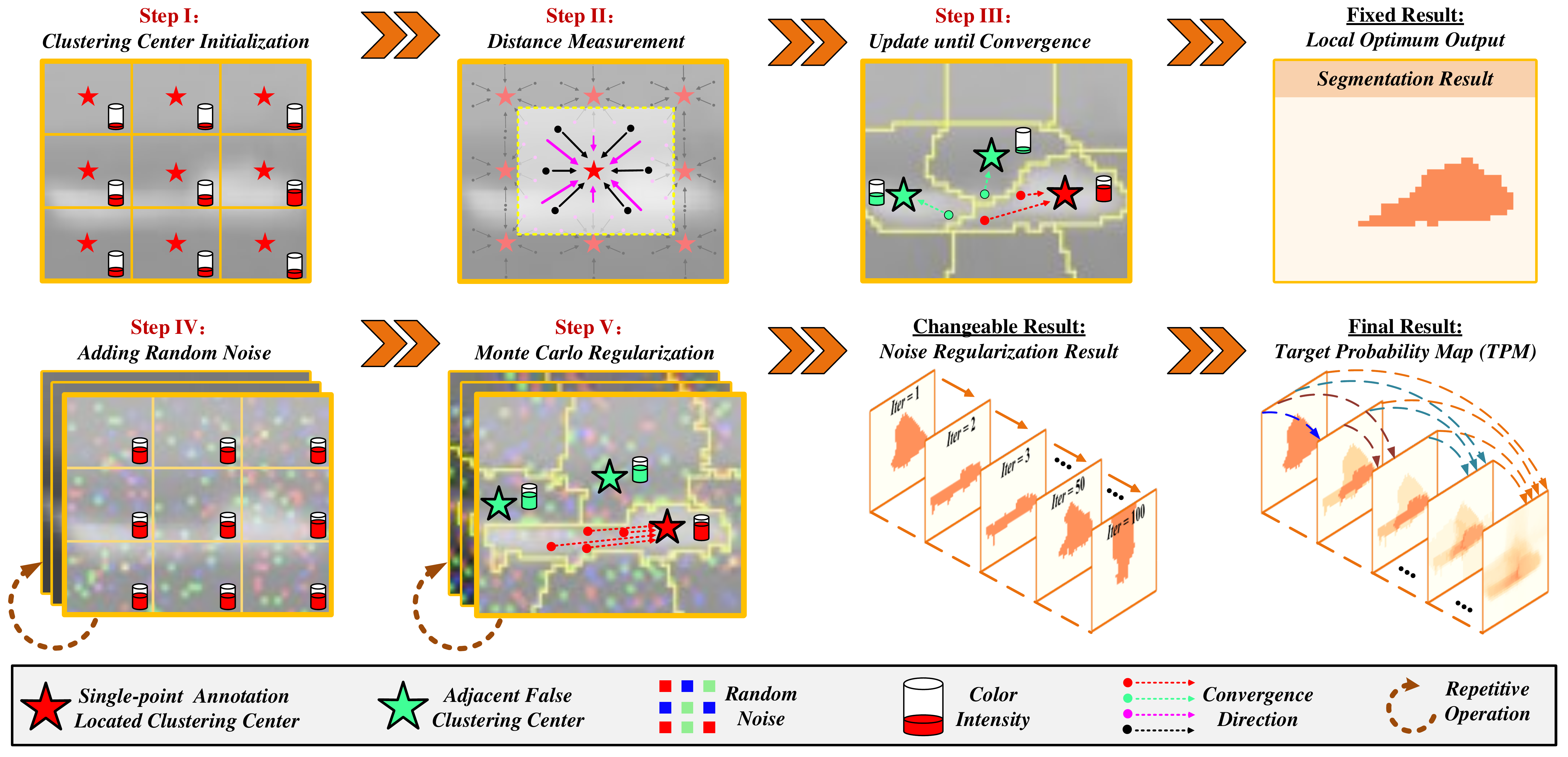}
\caption{Illustration of our method. $Step I$ $\sim$ $Step III$: LCA can coarsely separate target from clutter background, but easily falls into local optimum and produce incomplete pseudo mask. $Step IV$ $\sim$ $Step V$: Monte Carlo regularization can increase the color distance between background and target region and thus enforce the misclustered target region away from the false clustering center. After repetitive random experiments, the accumulated clustering results are finally shown as a target probability map (TPM), where pixels with larger values represent a higher probability of belonging to the target. }\label{Fig_3}
\end{figure*}

\subsection{Linear Clustering Approach}\label{Sec_LC}

Without loss of generality, we take the quick linear clustering method  SuperPixel \cite{SLIC} as an example to introduce how to use LCA to coarsely separate the target region. It should be noted that other linear clustering approaches such as KMeans \cite{kmeans} can be also used in our method, as demonstrated in Fig.~\ref{Fig_5}. The pipeline of our LCA is shown in Fig.~\ref{Fig_3} (\textit{Step I} $\sim$ \textit{Step III}), in which the clustering center of the image is first initialized, and then adjusted according to a color and spatial distance based measurement. Finally, the clustering center is continuously updated until the pre-defined convergence requirements are met.

\textbf{\textit{1) Clustering Center Initialization:}} We initialize the clustering center based on the pre-defined number of clusters \emph{N}. Given an input image $\textbf{\emph{I}} \in \mathbb{R}^{ H \times W}$, we first divide it into \emph{N} regions with equal areas and fixed grid interval $\emph{S}= \sqrt{\frac{H\times W}{N}}$ ( as shown in Fig.~\ref{Fig_3}-\textit{Step I}). After that, the clustering center is first initialized as the centroid of each region, and then moved to the lowest gradient position of its 3$\times$3 neighborhood. The clustering center of the $n^{th}$ clustering region $\mathcal{C}_{n}$  can be represented as a vector as:

\begin{equation}
\label{equation_1}
\mathcal{C}_{n} =\left[c_{n}, s_{n}  \right]^{T},
\end{equation} where $\left[c_{n}\right]^{T}$  represents the CIELAB color \cite{CIEColor} of $\mathcal{C}_{n}$, $\left[s_{n} \right]^{T}$ denotes the spatial position in image coordinate of $\mathcal{C}_{n}$.

\textbf{\textit{2) Distance Measurement:}} After clustering center initialization, we adopt a color and spatial distance based measurement to determine the clustering center of each pixel. As shown in Fig.~\ref{Fig_3}-\textit{Step II}, for each clustering center $\mathcal{C}_{n}$, we use $L_{2}$ distance to measure its distance with each pixel $\emph{p}_{i}= \left[c_{i}, s_{i} \right]^{T}$ in its adjacent $2S \times 2S$ regions.
Considering that the ranges of color value and spatial position value are inconsistent, we normalize the both distance values with normalization coefficient $\mu_{s}$ and  $\mu_{c}$.
$\mathcal{D}({\mathcal{C}_{n}, p_{i}}) $ is written as:

\begin{equation}
\begin{split}
\label{equation_2}
\mathcal{D}(\mathcal{C}_{n}, p_{i}) & = \sqrt{(\mathcal{D}_{c})^{2} + (\mathcal{D}_{s})^{2}} \\
&= \sqrt{\frac{\left(c_{p_{i}}-c_{\mathcal{C}_{n}}\right)^{2}}{\mu_{c}^{2}} + \frac{\left(s_{p_{i}}-s_{\mathcal{C}_{n}}\right)^{2}}{\mu_{s}^{2}}}.
\end{split}
\end{equation}

 Then, the clustering label \emph{l} for each pixel $\emph{p}_{i}$ can be determined as:

\begin{equation}
\label{equation_3}
l(p_{i})   = \rm {argmin}_{n\in \{1,2,3,...,N\}}\left[\mathcal{D}({\mathcal{C}_{n}, p_{i})}\right].
\end{equation}

\textbf{\textit{3) Updating and  Convergence:}} After determining the clustering center of each pixel at the current iteration, we repetitively update the position of each clustering center until convergence, as shown in Fig.~\ref{Fig_3}-\textit{Step III}. Specifically, we first calculate the clustering centers to be the mean vector $\left[c_{m}, s_{m}\right]^{T}$ of all the pixels $P= \{ p_{1}, p_{2},...p_{m}\}$  with the same clustering label $l({\mathcal{C}^{'}_{n}})$. The new clustering center $\mathcal{C}^{'}_{n}$ can be formulated as:

\begin{equation}
\label{equation_4}
\mathcal{C}^{'}_{n}(c_{n}, x_{n}, y_{n}) = \frac{1}{\left|P\right|} \sum_{p_{m}\in P}(c_{p_{m}}, s_{p_{m}}).
\end{equation}

After that, we adopt $\emph{L}_{2}$ norm to compute a distance $\mathcal{E} $ between new clustering center locations $\mathcal{C}^{'}_{n}$ and previous clustering center locations $\mathcal{C}_{n}$. The updating steps can be repeated iteratively until the error is smaller than the pre-defined threshold \emph{T}. That is $\Vert\mathcal{C}^{'}_{n} - \mathcal{C}_{n}\Vert_2  < \emph{T}$, where $\Vert \cdot \Vert_2$ denote the L2 norm. Finally, we assign the single-point annotation $p_{Anno}$ located at the $n^{th}$ clustering region as the clustering result $\mathbf{\emph{M}}_{pred}$.

\begin{figure}
\centering
\includegraphics[width=0.90\linewidth]{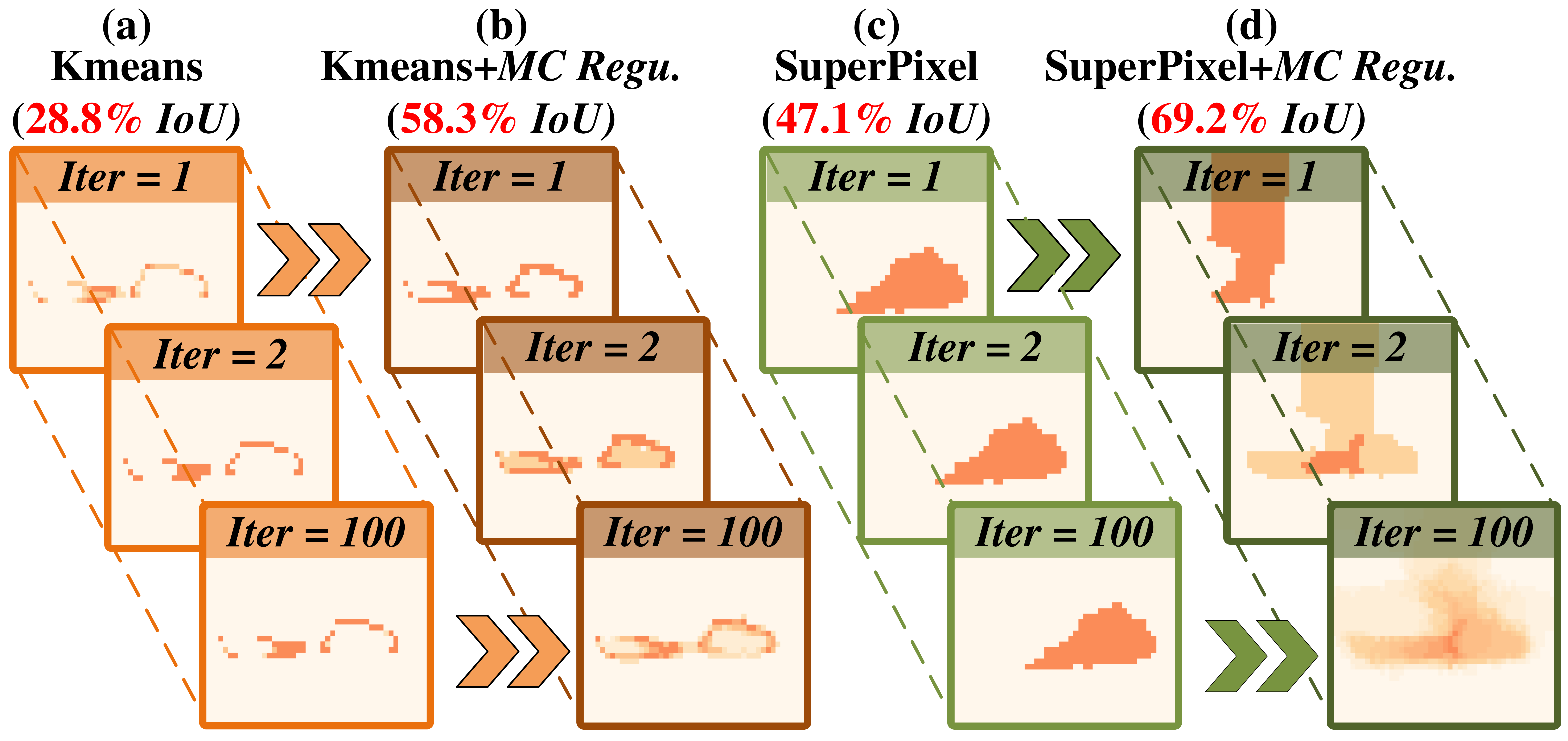}
\caption{Visualization results achieved by (a) Kmeans, (b) Kmeans with Monte Carlo Regularization, (c) Super pixel, (d) Super pixel with Monte Carlo Regularization. Our proposed Monte Carlo regularization helps to expand informative target region. }\label{Fig_5}
\end{figure}

\subsection{Monte Carlo Regularization}\label{Sec_optimization}

Although LCA can coarsely separate local salient regions from the backgrounds, as shown in Fig.~\ref{Fig_5} (a) and (c), a linear clustering method easily falls into local optimum and generates incomplete or oversized results. In this subsection, we first analyze the reason of this issue, and then propose a Monte Carlo regularization to alleviate this issue, as shown in Fig.~\ref{Fig_3} (\textit{StepIV} $\sim$ \textit{StepV}).

\textbf{\textit{1) Inaccurate Result within a Single Clustering:}} Due to the varied size of small targets, LCA with fixed number of clustering centers cannot produce ideal results for all kinds of targets. That is because, both color $\mathcal{D}_{c}$ and spatial distance $\mathcal{D}_{s}$ determine the output of linear clustering model in Equation~\ref{equation_2}. Since edge areas  $\mathbf{\emph{M}}^{Edge}_{pred} = \{ \mathbf{\emph{m}}^{Edge(1)}_{pred}, \mathbf{\emph{m}}^{Edge(2)}_{pred},...,\mathbf{\emph{m}}^{Edge(Z)}_{pred}\}$ of the spot and extended target (i.e., big target) are generally far from the true clustering center $\mathcal{C}^{T}$ (shown as the left-most green point in Fig.~\ref{Fig_3}-\textit{Step III}). The spatial distance $\mathcal{D}_{c}(\mathcal{C}^{T}, \mathbf{\emph{M}}^{Edge}_{pred})$ between the true clustering center $\mathcal{C}^{T}$ and edge areas $\mathbf{\emph{M}}^{Edge}_{pred}$ is usually father than that of the false clustering center $\mathcal{C}^{F}$. That is:

\begin{equation}
\label{equation_7}
 \mathcal{D}_{s}(\mathcal{C}^{T}, \mathbf{\emph{m}}^{Edge(z)}_{pred}) > \mathcal{D}_{s}(\mathcal{C}^{F},  \mathbf{\emph{m}}^{Edge(z)}_{pred}).
\end{equation}

\begin{figure}
\centering
\includegraphics[width=0.99\linewidth]{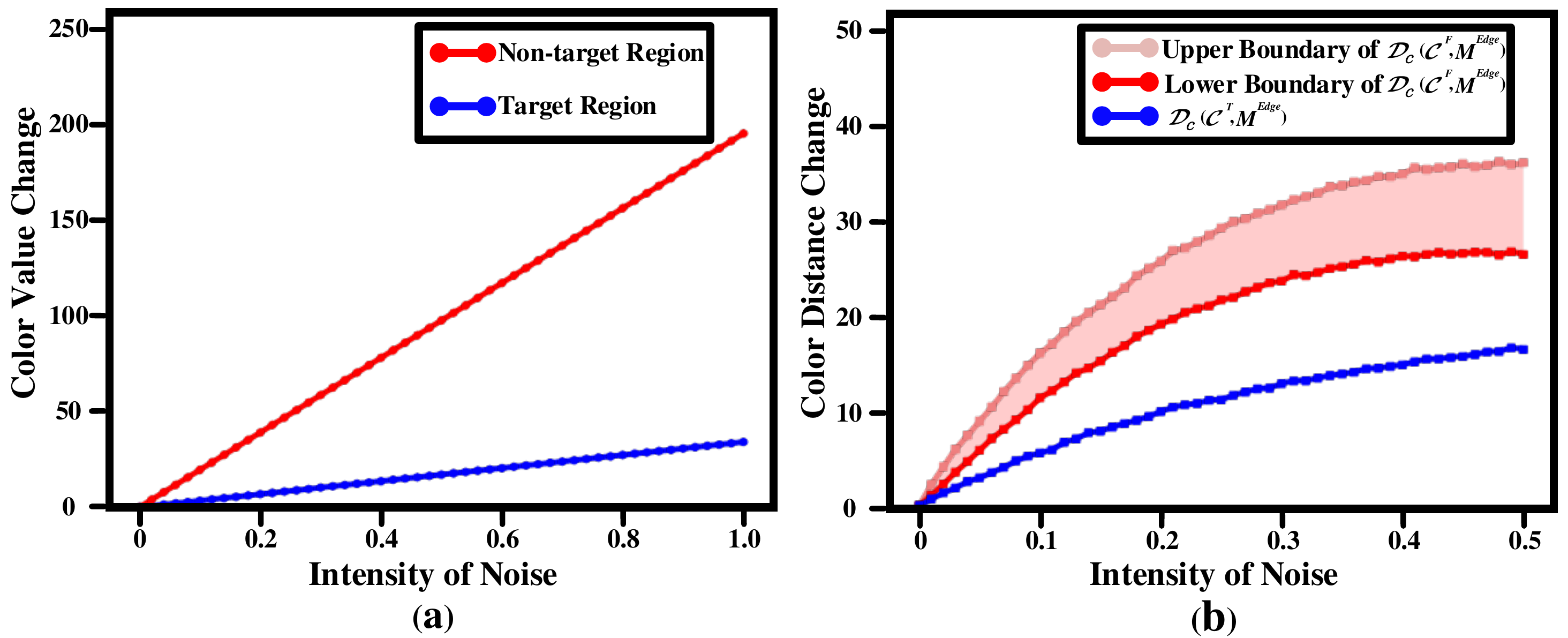}
\caption{Quantitative analysis on (a) Sensitivity of target and non-target regions to additional noise, and (b) Color distance of target edge region with true foreground region centroid $ \Delta (\mathcal{D}_{c}(\mathcal{C}^{T}, \mathbf{\emph{M}}_{pred}))$  and false background region centroid $ \Delta (\mathcal{D}_{c}(\mathcal{C}^{F}, \mathbf{\emph{M}}_{pred}))$. Average results from 10 trails are reported. Readers can refer to the supplementary material for more detailed illustration.}\label{Fig_10}
\end{figure}

Moreover, the color value of big target generally exhibits Gaussian distribution. Those edge areas often exhibit relatively high color difference with the true clustering center, and are close to the adjacent false clustering center (shown as the right-most green point in Fig.~\ref{Fig_3} \textit{Step III}). That is:

\begin{equation}
\label{equation_8}
 \mathcal{D}_{c}(\mathcal{C}^{T},\mathbf{\emph{m}}^{Edge(z)}_{pred}) > \mathcal{D}_{c}(\mathcal{C}^{F},  \mathbf{\emph{m}}^{Edge(z)}_{pred}).
\end{equation}

As a result, the label of edge areas $l^{Edge}_{pred}$ may be falsely included into the adjacent false clustering center $\mathcal{C}^{F}$, resulting in incomplete clustering result.

\textbf{\textit{2) Random Noise Guided Regularization:}} Spatial distance $\mathcal{D}_{s}$ is generally fixed and hard to change in the image. As shown in Fig.~\ref{Fig_10} (a), when adding random noise, target and non-target regions exhibit different sensitivity to the noise. Target regions with higher color vaule are more robust to the additional noise than the non-target regions. Based on this finding, we are motivated to introduce random noise to enlarge the color distance $ \mathcal{D}_{c}(\mathcal{C}^{F}, \mathbf{\emph{M}}^{Edge}_{pred})$ between edge areas and false clustering center (as shown in Fig.~\ref{Fig_3}-\textit{Step III}), and thus make the edge areas be clustered to the true clustering center. Random noise-regularized clustering result can be formulated as follows:

\begin{equation}
\label{equation_9}
\mathbf{\emph{M}}^{Noise}_{pred} = LCA(clip(\mathbf{\emph{I}}+ \mathcal{N})),
\end{equation} where $clip(\cdot)$ operation represents that color values larger than 255 will be fixed at 255. $\mathcal{N}$ is the additional random noise. The color value change caused by random noise can be formulated as:

\begin{equation}
\label{equation_10}
  \Delta (\mathcal{D}_{c}(\mathcal{C}, \mathbf{\emph{M}}_{pred})) = \mathcal{D}^{Noise}_{c} - \mathcal{D}^{Clean}_{c}.
\end{equation}

Since background region has relatively low color value than foreground region. Additional noise will greatly increase the color value of adjacent false clustering centers, and thus enlarges their color distance with the edge area of target region. On the contrary, when comparing Fig.~\ref{Fig_3}-\textit{Step IV} with Fig.~\ref{Fig_3}-\textit{Step V}, we observe that foreground regions have high energy concentricity and are usually close to saturation. Additional noise cannot bring so much color change to target regions as the background regions. Benefited by the additional noise, as shown in Fig.~\ref{Fig_10} (b), the change of color distance between edge areas and false clustering center (i.e., $ \Delta (\mathcal{D}_{c}(\mathcal{C}^{F}, \mathbf{\emph{M}}_{pred}))$) will be significantly larger than that between edge areas and true clustering center (i.e., $\Delta (\mathcal{D}_{c}(\mathcal{C}^{T}, \mathbf{\emph{M}}_{pred}))$), making the misclustered target region be more closed to the true clustering center.

\textbf{\textit{3) Monte Carlo Process:}}  As aforementioned, random noise can enforce the misclustered region to be gradually close to the true clustering center. Considering the randomness of noise, additional noise does not always bring such positive effect. We are motivated to adopt the Monte Carlo method to accumulate the clustering result from repetitive random experiments and gradually recover the reliable clustering result. To this end, we perform \emph{K} independent clustering and formulate this process as follows:

\begin{equation}
\label{equation_11}
\mathbf{\emph{M}}^{Full}_{Pred} =  \frac{1}{K}  \sum_{k = 1}^{K}\left(   LCA(clip(\mathbf{\emph{I}}+ \mathcal{N}^{(k)}))  \right) ,
\end{equation} where $\mathbf{\emph{M}}^{Full}_{Pred}$ is the accumulated clustering results after \emph{K} independent clustering by $LCA$, and finally shown as a target probability map (TPM), where pixels with larger values
represent a higher probability of belonging to the target.

The optimization process finally stops when the distance between the new clustering center locations $\mathcal{C}^{'}$ and previous clustering center locations $\mathcal{C}$ is smaller than the pre-defined threshold $\emph{T}^{F}$, i.e., $\Vert\mathcal{C}^{'}_{n} - \mathcal{C}_{n}\Vert_2 < \emph{T}^{F}$.

\section{Experiments}\label{SecExperiment}

In this section, we first introduce our evaluation metrics and implementation details. Then, we compare our MCLC to several state-of-the-art unsupervised, point-level supervised, and fully-supervised SIRST detection methods. Finally, we present ablation studies to investigate the effectiveness of our method. Note that, we search for optimal parameters of MCLC in the training set of NUAA-SIRST \cite{ACM} and directly adopt them as default parameters to generate pseudo masks on the other three datasets\cite{Shape-matter, DNANet, SIRST-sea}.

\subsection{Evaluation Metrics}\label{Evaluation Metrics}

Considering that both the quality of pseudo labels and the final detection results are crucial to SIRST detection, we followed previous works \cite{SEAM,CPN,PSA,IRN} and propose a two-stage pipeline (i.e., pseudo mask and final detection result evaluation) to comprehensively evaluate the effectiveness of the proposed method. Firstly, we followed the same metrics in weakly-supervised general object segmentation \cite{PSA,IRN} and adopted intersection of union ($IoU$) to evaluate the quality of generated pseudo labels on the training set. For the final detection results evaluation, we followed previous works \cite{ACM, ALCNet, DNANet} and adopted probability of detection (${P}_{d}$), and false-alarm rate ${F}_{a}$ to evaluate the localization precision, and used $IoU$ to evaluate the shape description ability on the test set.

\begin{figure}
\centering
\includegraphics[width=0.99\linewidth]{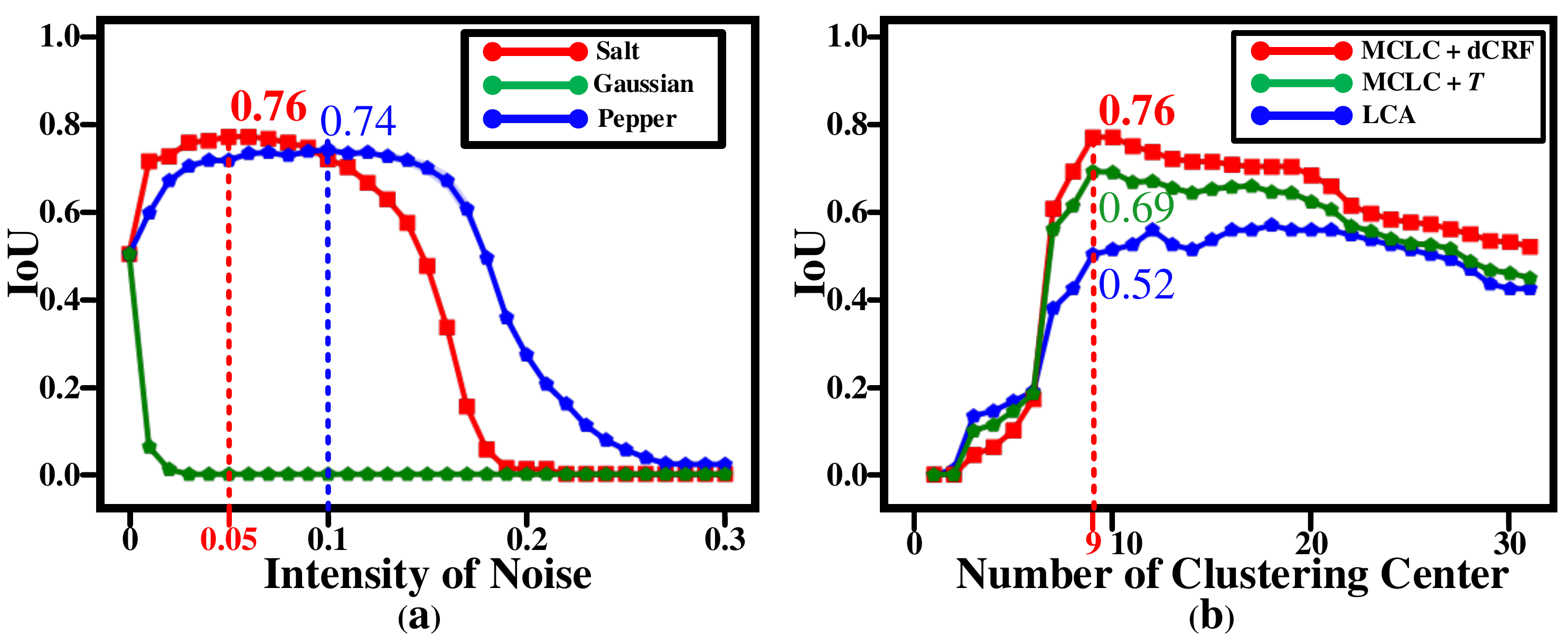}
\caption{ $IoU$ scores achieved by our method with (a) different types and intensity of additional noise and (b) different number of clustering center on the NUAA-SIRST dataset. Average results from 5 trails are reported. Readers can refer to the supplementary material for additional experiments on the other datasets.}\label{Fig_7}
\end{figure}

\begin{table}\scriptsize
\centering
\renewcommand\arraystretch{1.1}
\caption{$IoU(10^{-2})$ value achieved by the different variants of our method on the training set of four representative datasets (i.e., NUAA, IRSTD, NUDT, NUDT-sea). MC Regu. refers to Monte Carlo regularization.} \label{table1}
\begin{tabular}{c|c|c|c|c}
\hline
Point Label   & LCA               & MC Regu.      & dCRF           &  $IoU$(\%)                                 \\   \hline
$\checkmark$  &                   &               & $\checkmark$   &  2.53 $/$ 1.62  $/$ 2.28  $/$ 1.82         \\
$\checkmark$  & $\checkmark$      &               &                &  51.8 $/$ 55.5  $/$ 51.2  $/$ 38.1         \\
$\checkmark$  & $\checkmark$      & $\checkmark$  &                &  70.4 $/$ 66.2  $/$ 52.5  $/$ 43.9         \\
$\checkmark$  & $\checkmark$      & $\checkmark$  & $\checkmark$   &  \textbf{76.3} $/$ \textbf{68.8}  $/$ \textbf{60.2}  $/$ \textbf{47.1}         \\   \hline
\end{tabular}
\end{table}

\subsection{Implementation Details}\label{Protocol}


\textbf{\textit{1) Datasets:}} We evaluated our method on the NUAA-SIRST \cite{ACM}, IRSTD-1k \cite{Shape-matter}, NUDT-SIRST \cite{DNANet}, and NUDT-SIRST-sea \cite{SIRST-sea} datasets. For the NUAA-SIRST and NUDT-SIRST datasets, we used the same division as in \cite{DNANet} and set the ratio of train-val set to test set as 1. For the IRSTD-1k dataset, we followed \cite{Shape-matter} and used 80\% data for training and validation, and the remaining 20\% data for test. The division setting in \cite{SIRST-sea} was adopted for the NUDT-SIRST-sea dataset, i.e., 41 images were used for training and validation, and the remaining 7 images were used for test. Note that, only single point-level annotation with category information (i.e., point, spot, and extend categories) are available during network training.

\textbf{\textit{2) Training Details:}} In our experiments, all input images were first normalized before training. Then, these normalized images were sequentially processed by random flip, and random crop for data augmentation. Next, these images were resized to a fixed resolution. Specifically, 1024 $\times$1024 and 256 $\times$ 256 spatial resolutions were used for the NUDT-SIRST-sea and the remaining three datasets, respectively. Finally, we fed these resized datasets into network for both training and evaluation. Without specification, we adopted the centroid of ground truth mask as the single-point label. The influence of label position deviation is discussed in the following ablation. Moreover, all networks were trained with the Soft-IoU loss function and optimized by the Adagrad method \cite{Adagrad} with the CosineAnnealingLR scheduler.  The Xavier method \cite{xavier} was used for all weights and bias parameters initialization. We set the learning rate and epoch number to 0.05 and 1500, respectively. The batch size was set to 16 for the NUAA-SIRST, IRSTD-1k, NUDT-SIRST datasets but set to 4 for the NUDT-SIRST-sea dataset due to its large image size. All models were implemented in PyTorch \cite{Pytorch} on a computer with an AMD Ryzen 9 3950X @ 2.20 GHz CPU and an Nvidia GeForce 3090 GPU.

\subsection{Ablation Study}\label{Ablation}

\begin{figure}
\centering
\includegraphics[width=0.99\linewidth]{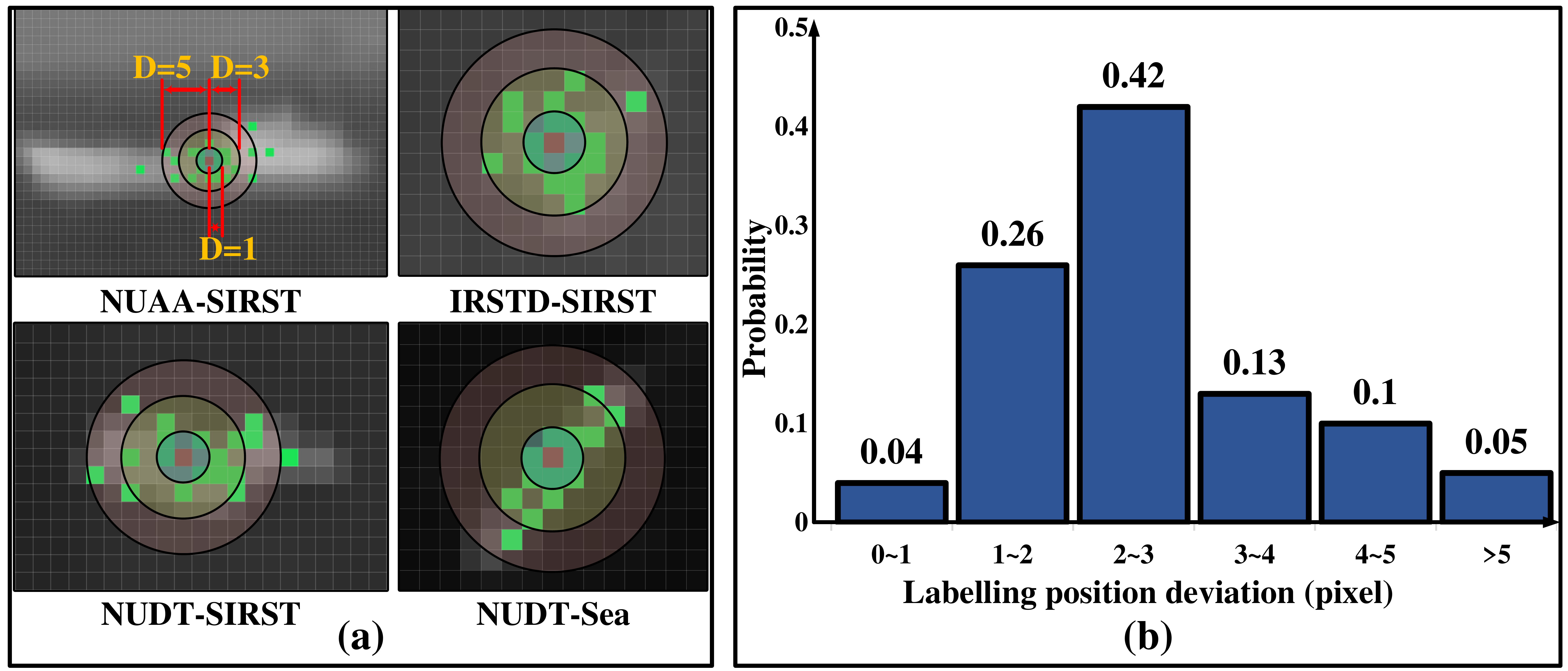}
\caption{(a) Labelling deviation examples of small target from four datasets. (b) Label deviation distribution map generated from 105 manually re-labelled targets. Real world labelling deviation of small target is generally less than 3 pixels.}\label{Fig_11}
\end{figure}

\textbf{\textit{1) Effectiveness of Proposed Components:}} Table~\ref{table1} shows the contribution of our proposed components on four datasets. Compared to the initial single-point label, linear clustering approach (LCA) introduces 49.3\%, 53.9\%, 48.9\%, 36.3\% improvements for the NUAA, IRSTD-1k, NUDT, NUDT-sea datasets, respectively. That is because, LCA utilizes the inherent characteristics of infrared small targets and thus coarsely separates the targets from clustering background. By introducing random salt noise to regularize the linear clustering process, we obtain much more reliable results and achieve additional 18.6\%, 10.7\%, 1.3\%, 5.8\% improvements on the above four datasets. After that, denseCRF \cite{deeplabv1} was used as a post-processing module to refine the target probability map (TPM) generated by MCLC. It further introduces 5.9\%, 2.6\%, 7.7\%, 3.2\% improvements for the above four datasets, respectively. We visualize the Monte Carlo linear clustering process in Fig.~\ref{Fig_8}. Although easily producing inaccurate results at the beginning of clustering (e.g., iteration number less than 20), MCLC can gradually recover a reliable clustering result. More visualization samples are shown in supplementary material.

\textbf{\textit{2) Type and Intensity of Noise:}} Fig.~\ref{Fig_7} (a) shows the change trend of $IoU$ with respect to different noise intensity under three types of common noise (i.e., salt, pepper, and Gaussian). With the increase of noise intensity, the $IoU$ of MCLC with denseCRF under salt noise and pepper noise increases rapidly at the beginning, and reaches their peak scores of 76\% and 71\% at 0.05 intensity\footnote{The intensity value for salt and pepper noise is defined as the area ratio of noise region and image region.  Higher values denote that more pixels are replaced by salt or pepper pixels.}. After that, excessive intensity value reduces the saliency of target region and thus results in the decrease of $IoU$. Moreover, Gaussian noise causes huge performance decrease under any intensity. That is because, Gaussian noise randomly changes the values of all pixels in the image, which is similar to the salt and pepper with high intensity. Since all pixels in the image are influenced by the random noise, the saliency of target is greatly decreased, resulting in dramatic performance drop.

\begin{table}\scriptsize
\centering
\renewcommand\arraystretch{1.1}
\caption{$IoU(10^{-2})$ values and corresponding labelling cost with stronger supervision on the train set of four datasets.} \label{table2}
\begin{tabular}{c|c|c|c|c}
\hline
\multirow{2}{*}{Annotation} & \multicolumn{4}{c}{Dataset}                                                                  \\ \cline{2-5}
                            & \multicolumn{1}{c|}{NUAA}     & \multicolumn{1}{c|}{IRSTD}     & \multicolumn{1}{c|}{NUDT}    & NUDT-sea \\ \hline
Single-Point                & \multicolumn{1}{c|}{76.3}     & \multicolumn{1}{c|}{68.8}      & \multicolumn{1}{c|}{60.2}    & \multicolumn{1}{c}{47.1}         \\
Multi-Point                 & \multicolumn{1}{c|}{76.6}     & \multicolumn{1}{c|}{69.1}      & \multicolumn{1}{c|}{62.6}    & \multicolumn{1}{c}{53.7}         \\
Scribble                    & \multicolumn{1}{c|}{78.5}     & \multicolumn{1}{c|}{70.5}      & \multicolumn{1}{c|}{63.2}    & \multicolumn{1}{c}{55.2}         \\
Bounding-Box                & \multicolumn{1}{c|}{78.9}     & \multicolumn{1}{c|}{71.8}      & \multicolumn{1}{c|}{64.8}    & \multicolumn{1}{c}{54.5}        \\ \hline
\end{tabular}
\end{table}

\textbf{\textit{3) Number of Clustering Center:}} The number of clustering center determines the initial area of each clustering region. Fig.~\ref{Fig_7} (b) reports the $IoU$ of the results generated by MCLC with denseCRF,  MCLC with fixed threshold, and LCA under different number of clustering center. It shows that the $IoU$ firstly shows a rapid increasing trend with the increase number of clustering center, and then reaches the peak score of 76\% with 9 clustering centers. Afterwards, the quality of TPM gradually decreases when the number of clustering center further increases. The above results demonstrate that inappropriate number of clustering center will result in over-small or over-large area of initial search region, and thus introduce negative effect on MCLC.

\textbf{\textit{4) Performance with Stronger Supervision:}} As shown in Table~\ref{table2}, stronger supervision can introduce additional 0.3\% $\sim$ 2.6\% improvements in term of $IoU$ on the NUAA-SIRST dataset, but at a cost of 121\% $\sim$ 700\% increase of annotation time (i.e., 1.4s, 3.1s, and 11.2s for single-point, multi-point, and pixel-level annotation in Fig.~\ref{Fig_4}). Similar results can be also found on the other three datasets. Therefore, we argue that single-point annotation is the most economical supervision and is strong enough for SIRST detection.

\begin{table}[]\scriptsize
\centering
\renewcommand\arraystretch{1.1}
\caption{$IoU(10^{-2})$ values achieved with single point supervision under ideal and real scene.} \label{table4}
\begin{tabular}{l|c|c|c|c}
\hline
        & \begin{tabular}[c]{@{}c@{}}Pseudo Mask \\ (Ideal / Real)\end{tabular} & \begin{tabular}[c]{@{}c@{}}Final Results \\ (Ideal / Real)\end{tabular}
        & \begin{tabular}[c]{@{}c@{}}Pseudo Mask \\ (Ideal / Real)\end{tabular} & \begin{tabular}[c]{@{}c@{}}Final Results \\ (Ideal / Real)\end{tabular}\\ \hline \hline
ResUnet   &  (76.3 / 73.7)    & (71.6 / 71.2)          & (60.2 / 58.8)           & (68.0 / 66.5)                         \\ \cline{1-1} \cline{3-3} \cline{5-5}
DNANet    &  \textbf{NUAA}    & (72.9 / 72.0)          & \textbf{NUDT}           & (70.5 / 69.1)                         \\ \hline
ResUnet   &  (68.8 / 66.7)    & (64.6 / 63.8)          & (47.1 / 45.7)           & (39.1 / 38.2)                         \\ \cline{1-1} \cline{3-3} \cline{5-5}
DNANet    &  \textbf{IRSTD}   & (62.2 / 60.9)          & \textbf{NUDT-Sea}       & (40.2 / 38.8)                         \\ \hline
\end{tabular}
\end{table}

\begin{table}
\centering
\scriptsize
\caption{$IoU(10^{-2})$ values under fixed labelling time. 5.2$\times$ (with search) and 8.0$\times$ (w/o search) more point-level (P.) labels help to generate better results under fixed labelling time.} \label{table3}
\begin{tabular}{l|c|ccc}
\hline
\multirow{2}{*}{}                 & \multirow{4}{*}{Sup.} & \multicolumn{1}{c|}{NUAA-SIRST}                                         & \multicolumn{1}{c}{IRSTD-1k}                       \\ \cline{3-4}
                                  &                       & \multicolumn{1}{c|}{Time Budget: 920s}                                  & \multicolumn{1}{c}{Time Budget: 1400s} \\
                                  &                       & \multicolumn{1}{c|}{(427$\times$\textit{P.}) vs (82$\times$\textit{F.})}  & \multicolumn{1}{c}{(1000$\times$\textit{P.}) vs (125$\times$\textit{F.})}  \\
                                  &                       & (5.2$\times$ P. labels)           & (8.0$\times$ P. labels)  \\ \hline
ResUnet                                    & \textit{Full}        & \multicolumn{1}{c|}{56.8}                                       & \multicolumn{1}{c}{43.8}               \\ \hline
\textbf{ResUnet}+\textit{\textbf{MCLC}}    & \textit{P.}       & \multicolumn{1}{c|}{\textbf{71.6}  $\textcolor{red}{\uparrow 14.8}$}     & \multicolumn{1}{c}{\textbf{64.6}  $\textcolor{red}{\uparrow 20.8}$}          \\ \hline
DNANet                                     & \textit{Full}        & \multicolumn{1}{c|}{58.7}                                       & \multicolumn{1}{c}{46.1}              \\ \hline
\textbf{DNANet}+\textit{\textbf{MCLC}}     & \textit{P.}       & \multicolumn{1}{c|}{\textbf{72.1}  $\textcolor{red}{\uparrow 13.4}$}     & \multicolumn{1}{c}{\textbf{62.2} $\textcolor{red}{\uparrow 16.1}$}              \\ \hline
\end{tabular}
\end{table}

\begin{table*}[]\scriptsize
\centering
\renewcommand\arraystretch{0.95}
\caption{$IoU(10^{-2})$, $P_{d}(10^{-2})$, and $F_{a}(10^{-6})$ values achieved by different state-of-the-art methods on four benchmark datasets. For $IoU$ and $P_{d}$, larger values indicate better performance. For $F_{a}$, smaller values indicate better performance. \textit{Unsup.} refers to unsupervised methods. The best single-point supervised results are in \textcolor{red} {red} and the second best results are in \textcolor{blue} {blue}.} \label{table5}
\begin{tabular}{l|c|c|c|c|c|c|c|c|c|c|c|c|c}
\hline
\multirow{2}{*}{Method}             & \multirow{2}{*}{Sup.}        & \multicolumn{12}{c}{Dataset}                      \\  \cline{3-14}
                                    &                              & \multicolumn{3}{c|}{NUAA-SIRST \cite{ACM}} & \multicolumn{3}{c|}{IRSTD-1k \cite{Shape-matter}} & \multicolumn{3}{c|}{NUDT-SIRST \cite{DNANet}} & \multicolumn{3}{c}{NUDT-SIRST-sea \cite{SIRST-sea}}          \\ \hline \hline
Top-Hat      \cite{4-tophat}        & $\textit{Unsup.}$                           & 7.143& 79.84& 1012   & 6.222& 55.48& 595.6    & 20.72& 78.41& 166.7  & 1.17& 2.680& 95.37      \\
WSLCM        \cite{9-WSLLCM}        & $\textit{Unsup.}$                           & 1.158& 77.95& 5446   & 0.019& 55.79& 31547   & 2.283& 56.82& 1309   & 0.60& 10.52& 7.330         \\
MSLSTIPT     \cite{10-IPI}          & $\textit{Unsup.}$                           & 10.30& 82.13& 1131   & 10.37& 57.05& 3707    & 8.342& 47.40& 888.1  & 0.33& 0.350& 6283         \\
NRAM         \cite{11-NRAM}         & $\textit{Unsup.}$                           & 12.16& 74.52& 13.85  & 4.221& 49.21& 27.86   & 6.927& 56.40& 19.27  & 0.35& 17.31& 3.585         \\
PSTNN        \cite{13-PSTNN}        & $\textit{Unsup.}$                           & 22.40& 77.95& 29.11  & 9.871& 51.72& 99.75   & 14.85& 66.13& 44.17  & 1.50& 13.51& 15.44        \\ \hline

ACM          \cite{ACM}             & $\textit{Full}$                                  & 70.33& 93.91& 3.728                                                     & 60.97& 90.58& 21.78
                                                                                       & 67.08& 95.97& 10.18                                                     & 47.57& 70.46& 21.31        \\
\rowcolor{gray!15} ACM      + \textit{KMeans}\cite{kmeans}   & $\textit{Point}$      & 46.16& 88.59& \textcolor{blue}{27.45}                                   & 33.41& 54.42& 41.82
                                                                                       & 41.18& 77.88& 34.86                                                     & 38.15& 42.45& 23.25        \\
\rowcolor{gray!15} ACM      + \textit{SuperPixel}\cite{SLIC}  & $\textit{Point}$     & 53.84& 90.49& 56.18                                                     & 49.69& \textcolor{red}{84.69}& 42.36
                                                                                       & 51.76& \textcolor{blue}{92.33}& 34.60                                   & \textcolor{blue}{40.42}& 36.70& 16.92        \\
\rowcolor{gray!15} ACM      + \textit{GrabCut}\cite{Grabcut}  & $\textit{Point}$     & \textcolor{blue}{57.63}& \textcolor{blue}{90.87}                        & 60.46& \textcolor{blue}{54.03}& 81.97
                                                                                       & \textcolor{red}{23.99} & 55.24& 91.65& \textcolor{red}{6.457}           & 38.76& \textcolor{blue}{52.37}& \textcolor{blue}{15.54}        \\
\rowcolor{gray!15} ACM      + \textit{\textbf{Ours}}  & $\textit{Point}$       & \textcolor{red}{67.08}& \textcolor{red}{92.01}& \textcolor{red}{19.80}  & \textcolor{red}{56.53}& \textcolor{blue}{82.31}& \textcolor{blue}{25.05}
                                                                                 & \textcolor{red}{57.74}& \textcolor{red}{92.38}& \textcolor{blue}{16.77} & \textcolor{red}{43.69}& \textcolor{red}{55.69}& \textcolor{red}{9.820}       \\ \hdashline

ResUnet\cite{SIRST-sea}       & $\textit{Full}$                                        & 75.93& 97.71& 15.68                                                      & 66.34& 92.83& 8.198
                                                                                       & 83.74& 98.09& 4.820                                                      & 46.05& 60.18& 7.920        \\
\rowcolor{gray!15} ResUnet  + \textit{KMeans}\cite{kmeans}   & $\textit{Point}$      & 22.34& 39.92& 156.5                                                      & 33.86& 80.95& 10.32
                                                                                       & 42.65& 89.20& 42.51                                                      & 31.11& 51.74& 44.63        \\
\rowcolor{gray!15} ResUnet  + \textit{SuperPixel}\cite{SLIC}  & $\textit{Point}$     & 60.02& 93.15& \textcolor{blue}{23.45}                                    & 59.28& 88.43& 24.36
                                                                                       & 62.65& 93.40& 26.95                                                      & 36.43& \textcolor{blue}{56.50}& 30.61        \\
\rowcolor{gray!15} ResUnet  + \textit{GrabCut}\cite{Grabcut}  & $\textit{Point}$     & \textcolor{blue}{61.38}& \textcolor{red}{95.43}& 25.16                   & 61.67& 90.13& 13.96
                                                                                       & 63.01&  \textcolor{red}{94.03}& \textcolor{red}{13.44}                   & \textcolor{blue}{37.04}& 56.19&  \textcolor{blue}{17.32}     \\
\rowcolor{gray!15} ResUnet  + \textit{\textbf{Ours}}  & $\textit{Point}$       & \textcolor{red}{71.58}& \textcolor{blue}{94.67}& \textcolor{red}{15.21}  & \textcolor{red}{64.59}& \textcolor{red}{90.81}& \textcolor{red}{6.223}
                                                                                 & \textcolor{red}{68.04}&  \textcolor{blue}{93.86}& \textcolor{blue}{25.35}  & \textcolor{red}{39.06}& \textcolor{red}{59.44}&
                                                                                 \textcolor{red}{6.979}        \\ \hdashline

DNANet       \cite{DNANet}          & $\textit{Full}$                                  & 76.24& 97.71& 12.80                                                    & 68.44& 94.77& 8.806
                                                                                       & 86.36& 97.39& 6.897                                                    & 42.17& 61.60& 17.19        \\
\rowcolor{gray!15} DNANet   + \textit{KMeans}\cite{kmeans}   & $\textit{Point}$      & 45.69& 90.49& 58.11                                                      & 32.98& 81.63& 25.18
                                                                                      & 42.84& 88.04& 56.99                                                    & 20.59& 32.67& 21.59        \\
\rowcolor{gray!15} DNANet   + \textit{SuperPixel}\cite{SLIC}  & $\textit{Point}$     & \textcolor{blue}{62.59}& 93.53& \textcolor{blue}{14.54}                  & 58.44& 89.45& 27.70
                                                                                       & \textcolor{blue}{64.68}& 95.44& \textcolor{blue}{33.39}                & \textcolor{blue}{38.42}& 36.70& \textcolor{blue}{6.921}        \\
\rowcolor{gray!15} DNANet   + \textit{GrabCut}\cite{Grabcut}  & $\textit{Point}$     & 61.14& \textcolor{red}{97.71}& 20.53                            & \textcolor{blue}{61.00}& \textcolor{blue}{91.15}& \textcolor{red}{20.87}
                                                                                       & 64.00& \textcolor{red}{97.03}& 40.65                          & 28.08& \textcolor{blue}{49.79} & \textcolor{red}{4.816}        \\
\rowcolor{gray!15} DNANet   + \textit{\textbf{Ours}}  & $\textit{Point}$   & \textcolor{red}{72.86}& \textcolor{blue}{96.95}& \textcolor{red}{14.43}   &  \textcolor{red}{62.23}& \textcolor{red}{92.13}& \textcolor{blue}{24.14}
                                                                                       & \textcolor{red}{70.52}& \textcolor{blue}{95.55}& \textcolor{red}{33.20}   & \textcolor{red}{40.23}& \textcolor{red}{58.32}& 11.29        \\ \hdashline

ISNet       \cite{Shape-matter}          & $\textit{Full}$                             & 80.01& 99.23& 4.96                                                     & 68.72 & 95.68 & 15.43
                                                                                       & 82.57& 96.49& 44.11                                                    & 41.27 & 58.89 & 13.26        \\
\rowcolor{gray!15} ISNet   + \textit{KMeans}\cite{kmeans}   & $\textit{Point}$         & 38.27& 74.88& 112.7                                                    & 29.67 & 69.52 & 38.43
                                                                                       & 44.57& 89.87& 66.20                                                    & 32.57 & 44.20 & 35.26        \\
\rowcolor{gray!15} ISNet   + \textit{SuperPixel}\cite{SLIC}  & $\textit{Point}$        & \textcolor{blue}{65.82}& 93.73& 28.26                                  & 57.93 & 90.75 & 41.74
                                                                                       & 61.59& \textcolor{blue}{95.61}& \textcolor{red}{29.59}                 & \textcolor{blue}{35.21} & \textcolor{blue}{48.56} & \textcolor{red}{15.32}        \\
\rowcolor{gray!15} ISNet   + \textit{GrabCut}\cite{Grabcut}  & $\textit{Point}$        & 63.72& \textcolor{red}{96.27}& \textcolor{blue}{26.32}         & \textcolor{blue}{62.29} & \textcolor{blue}{91.37} & \textcolor{red}{19.34}
                                                                                       & \textcolor{blue}{65.53}& 94.89& \textcolor{blue}{45.31}                & 33.68 & 48.24& 23.56     \\

\rowcolor{gray!15} ISNet   + \textit{\textbf{Ours}}  & $\textit{Point}$      & \textcolor{red}{75.93} & \textcolor{blue}{95.44}& \textcolor{red}{11.04}   & \textcolor{red}{63.19}& \textcolor{red}{92.58}& \textcolor{blue}{22.23}
                                                                             & \textcolor{red}{66.87} & \textcolor{red}{96.22} & 48.59   & \textcolor{red}{38.29} & \textcolor{red}{52.27}& \textcolor{blue}{19.97}  \\ \hdashline

UIUNet       \cite{Shape-matter}          & $\textit{Full}$                             & 76.83& 97.64 & 12.31                                                      & 65.27& 92.76& 12.40
                                                                                        & 75.35& 93.33 & 27.69                                                      & 41.88& 55.69& 11.56        \\
\rowcolor{gray!15} UIUNet   + \textit{KMeans}\cite{kmeans}   & $\textit{Point}$         & 35.17& 90.90 & 46.53                                                      & 21.26& 82.52& 19.43
                                                                                        & 38.27& 88.27 & 78.09                                                      & 29.33& 33.17& \textcolor{red}{14.22}        \\
\rowcolor{gray!15} UIUNet   + \textit{SuperPixel}\cite{SLIC}  & $\textit{Point}$        & \textcolor{blue}{64.23}& 92.11 & 49.34                                    & 54.42& 87.30& 29.82
                                                                                        & 51.68& \textcolor{blue}{91.32} & 39.40                                 & \textcolor{blue}{37.83}& 47.22& \textcolor{blue}{20.09}        \\
\rowcolor{gray!15} UIUNet   + \textit{GrabCut}\cite{Grabcut}  & $\textit{Point}$        & 63.29& \textcolor{blue}{93.86} & \textcolor{blue}{33.27}        & \textcolor{blue}{60.37}& \textcolor{blue}{89.27}& \textcolor{red}{14.24}
                                                                                        & \textcolor{blue}{64.33}& \textcolor{red}{92.57} & \textcolor{red}{33.59}  & 37.03& \textcolor{blue}{51.55}& 26.12        \\
\rowcolor{gray!15} UIUNet   + \textit{\textbf{Ours}}  & $\textit{Point}$                & \textcolor{red}{74.22}& \textcolor{red}{96.20}& \textcolor{red}{16.19}    & \textcolor{red}{64.13}& \textcolor{red}{90.74}& \textcolor{blue}{14.93}
                                                                                        & \textcolor{red}{69.35}& 91.11 & \textcolor{blue}{38.21}            & \textcolor{red}{39.27}& \textcolor{red}{53.23}& 23.14       \\ \hline

\end{tabular}
\end{table*}

\textbf{\textit{5) Labelling Position Deviation:}} Real-world SIRST application may suffer from labelling position deviation shown in Fig.~\ref{Fig_11} (a). To simulate the labelling position deviation in real world, we manually re-labelled 105 targets in NUAA-SIRST and IRSTD-1k dataset, and generated a distribution map of labelling position in Fig.~\ref{Fig_11} (b). Based on this distribution map, we re-produced the pseudo masks of whole dataset and generated the final detection results in Table~\ref{table4}. Average results from three trails demonstrate that our method can produce competitive results under practical labelling settings.

\textbf{\textit{6) Time Efficiency:}} As aforementioned, once the optimal parameters (e.g., type and intensity of noise, number of clustering centers) are searched in one real-world dataset (e.g., NUAA-SIRST), we directly adopt these parameters to new scenes without secondary search (e.g., IRSTD in Table~\ref{table3}). As shown in Table~\ref{table3}, single-point labelling can produce about 5.2$\times$ (with parameter search) $\sim$ 8.0$\times$ (w/o parameter search) annotation than the pixel-level labelling with the same annotation budget, and thus achieves much better performance. Note that, our MCLC can efficiently recover a single point annotation to a pixel-level one with only 0.075s (average result from 533 targets) on a PC-level CPU.

\subsection{Comparison to the State-of-the-art Methods}\label{SOTA}

We compare our method to several state-of-the-art methods on four benchmark datasets in this subsection. For fair comparison, we used the same hyper-parameters as reported in their original papers when re-implementing comparing methods and retrained all the models from scratch on these four datasets.

\begin{figure}
\centering
\includegraphics[width=0.95\linewidth]{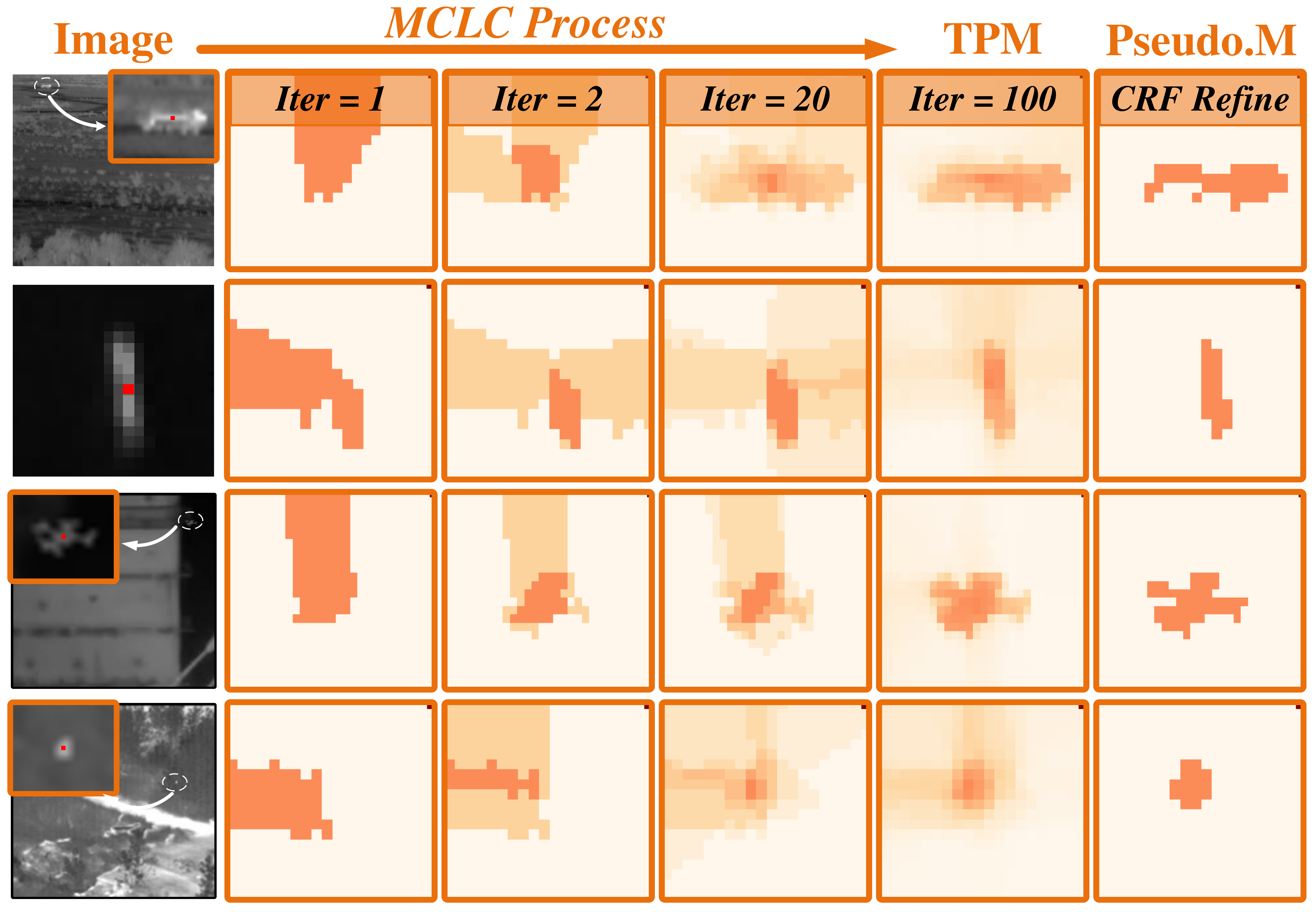}
\caption{Examples of target probability map (TPM) and the corresponding refined pseudo masks during the MCLC process.}\label{Fig_8}
\end{figure}

\begin{figure}
\centering
\includegraphics[width=0.95\linewidth]{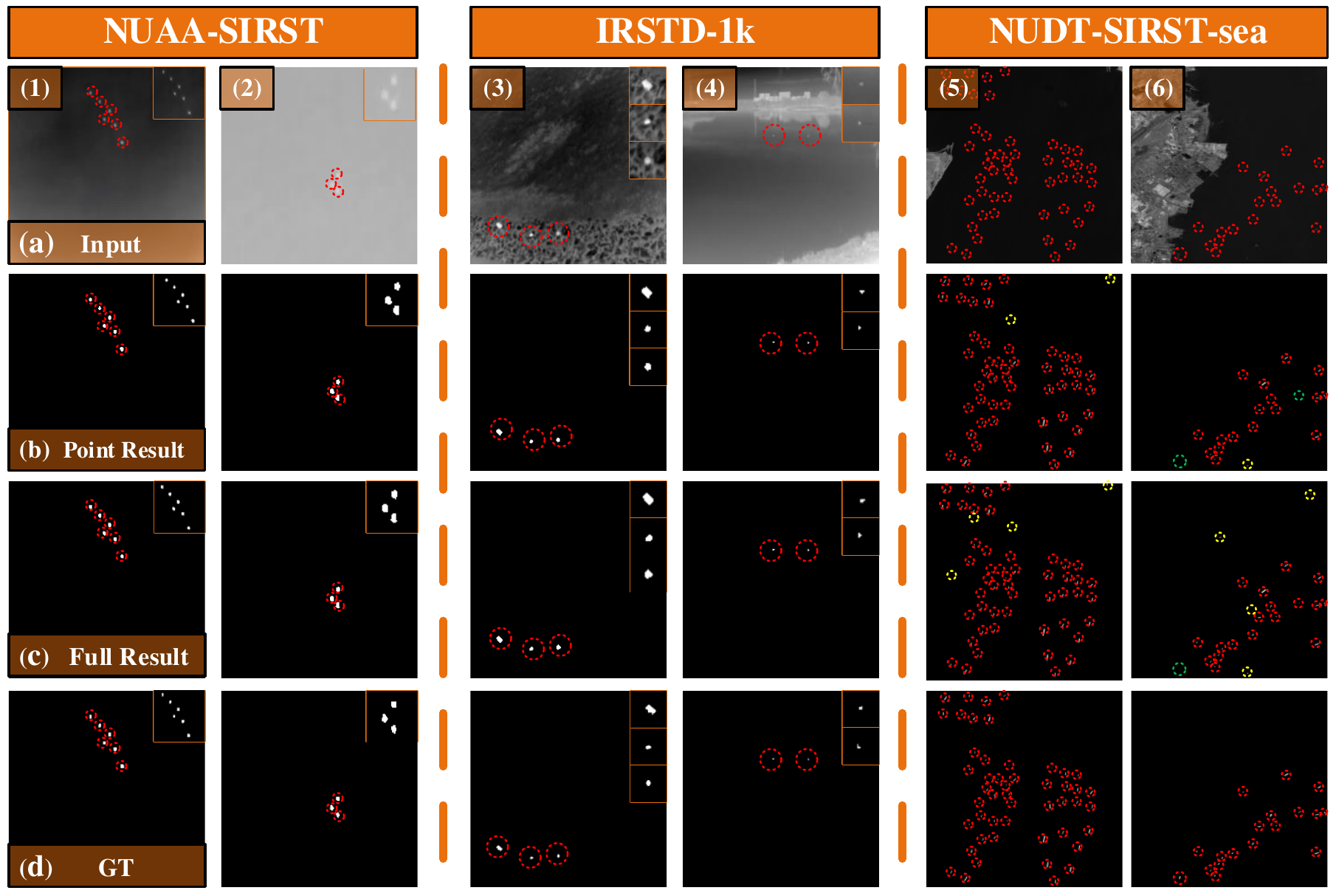}
\caption{Qualitative results achieved by DNANet\cite{DNANet} under (b) point-level and (c) pixel-level supervision. The correctly detected target, false alarm, and miss detection areas are highlighted by red, yellow, and green dotted circles. More visualization results are shown in supplementary material. }\label{Fig_9}
\end{figure}

\textbf{\textit{1) Quantitative and  Qualitative Results:}} Quantitative results are shown in Table~\ref{table5}. Our Monte Carlo linear clustering method (i.e., MCLC) achieves much better performance over those unsupervised methods. Compared to MCLC with other single-point supervised methods, MCLC can produce better pseudo masks and thus achieve better performance in term of ${P}_{d}$, ${F}_{a}$, and $IoU$. Compared to the fully-supervised counterparts, MCLC enables the respective networks to generate comparable result with 87\% annotation time reduction. Qualitative results on four datasets are shown in Fig.~\ref{Fig_9}. Compared to fully supervised methods, our MCLC can help existing detection methods to produce comparable results in a more time-efficient manner, especially for those point targets (e.g., img-1, img-4) and spot targets (e.g., img-2, img-3). Moreover, Fig.~\ref{Fig_9} (img-5. img-6) demonstrates the robustness of our method on the dense target scenarios. Readers can refer to the supplementary material for more visualization results.

\begin{table}[h]
\centering
\scriptsize
\caption{$IoU(10^{-2})$, $Pd(10^{-2})$, and $Fa(10^{-6})$ values achieved by existing salient object detection (SOD) methods.} \label{table6}
\begin{tabular}{l|c|ccc|ccc}\hline
                                  & \multicolumn{1}{c|}{Sup.} & \multicolumn{3}{c|}{NUAA-SIRST}                                                      & \multicolumn{3}{c}{IRSTD-1k}                    \\ \hline
ResUnet                                      & \textit{Full} & \multicolumn{1}{c|}{75.9} & \multicolumn{1}{c|}{97.7} & \multicolumn{1}{c|}{15.7} & \multicolumn{1}{c|}{66.3} & \multicolumn{1}{c|}{92.8} & \multicolumn{1}{c}{8.19}  \\ \hline
\rowcolor{gray!15}Res.+PFAN \cite{PFAN}   & \textit{Point} & \multicolumn{1}{c|}{27.6} & \multicolumn{1}{c|}{83.9} & \multicolumn{1}{c|}{49.6} & \multicolumn{1}{c|}{22.8} & \multicolumn{1}{c|}{76.5} & \multicolumn{1}{c}{66.2}  \\ \hline
\rowcolor{gray!15}Res.+F3Net \cite{F3Net} & \textit{Point} & \multicolumn{1}{c|}{33.2} & \multicolumn{1}{c|}{81.6} & \multicolumn{1}{c|}{106.2} & \multicolumn{1}{c|}{26.3} & \multicolumn{1}{c|}{71.3} & \multicolumn{1}{c}{82.3}  \\ \hline
\rowcolor{gray!15}\textbf{ResUnet}+\textit{\textbf{MCLC}} & \textit{Point} & \multicolumn{1}{c|}{\textbf{71.6}} & \multicolumn{1}{c|}{\textbf{94.7}} & \multicolumn{1}{c|}{\textbf{15.2}} & \multicolumn{1}{c|}{\textbf{64.6}} & \multicolumn{1}{c|}{\textbf{90.8}} & \multicolumn{1}{c}{\textbf{6.22}}  \\ \hdashline
DNANet                                & \textit{Full} & \multicolumn{1}{c|}{76.2} & \multicolumn{1}{c|}{97.7} & \multicolumn{1}{c|}{12.8} & \multicolumn{1}{c|}{68.4} & \multicolumn{1}{c|}{94.7} & \multicolumn{1}{c}{8.8} \\ \hline
\rowcolor{gray!15}DNA.+PFAN \cite{PFAN}   & \textit{Point} & \multicolumn{1}{c|}{29.3} & \multicolumn{1}{c|}{85.2} & \multicolumn{1}{c|}{66.8} & \multicolumn{1}{c|}{37.6} & \multicolumn{1}{c|}{78.6} & \multicolumn{1}{c}{57.7} \\ \hline
\rowcolor{gray!15}DNA.+F3Net \cite{F3Net} & \textit{Point} & \multicolumn{1}{c|}{33.6} & \multicolumn{1}{c|}{84.9} & \multicolumn{1}{c|}{78.0} & \multicolumn{1}{c|}{31.2} & \multicolumn{1}{c|}{77.1} & \multicolumn{1}{c}{61.2} \\ \hline
\rowcolor{gray!15}\textbf{DNANet}+\textit{\textbf{MCLC}} & \textit{Point}  & \multicolumn{1}{c|}{\textbf{72.9}} & \multicolumn{1}{c|}{\textbf{96.9}} & \multicolumn{1}{c|}{\textbf{14.4}} & \multicolumn{1}{c|}{\textbf{62.2}} & \multicolumn{1}{c|}{\textbf{92.1}} & \multicolumn{1}{c}{\textbf{24.1}} \\ \hline
\end{tabular}
\end{table}

\textbf{\textit{2) Compared with SOD Methods:}} Quantitative results in Table~\ref{table6} show that our MCLC performs much better than those salient object detection (SOD) methods. That is because, although the small targets are salient in local region, a single point label can not provide sufficient supervision to train high-performance SOD methods (e.g., PFAN \cite{PFAN}, F3Net \cite{F3Net}). In contrast, our MCLC can make full use of single point labels to obtain higher accuracy.

\section{Conclusion}

In this paper, we propose a simple yet effective pipeline for single-frame infrared small target detection with single-point annotation. First, we found that SIRST is generally salient in the local small region and exhibits high energy concentricity. Then, based on this observation, we proposed a linear clustering approach (LCA) to coarsely separate the targets from clustering background by measuring the color and spatial distance with clutter background. To further refine the results, we design a Monte Carlo regularization approach to constrain the clustering process and continuously expand the the point-level annotation to a high-quality pixel-level one. Extensive experiments on four benchmark datasets demonstrate that our method achieves comparable performance with fully-supervised methods. We hope our study can draw attention to the research on weakly supervised SIRST detection.

{\small
\bibliographystyle{ieee_fullname}
\bibliography{egbib}
}

\clearpage
\setcounter{section}{0}
\setcounter{figure}{0}
\setcounter{table}{0}

\renewcommand\thesection{\Alph{section}} 
\renewcommand\thetable{\Roman{table}}
\renewcommand\thefigure{\Roman{figure}}

\title{\textbf{Monte Carlo Linear Clustering with Single-Point Supervision is Enough for Infrared Small Target Detection \\ ({\textit{Supplemental Material}}) \vspace{-3em}}}

\author{}

\date{}
\maketitle

Section~\ref{sec:details_of_annotation} introduces the details of the annotation cost experiment. Section~\ref{sec:details_of_Regularization} discusses the results of the color distance change experiment. Section~\ref{sec:details_of_TPM} provides more target probability maps (TPMs) on different SIRST datasets. Section~\ref{sec:details_of_ablation} presents additional ablation studies on the other three SIRST datasets. More quantitative results on different SIRST datasets are shown in Section~\ref{sec:details_of_quantitative}. Moreover, we developed an offline webpage to summarize the pipeline, methods, and visual results of our paper. Readers can refer to the attached files for more details.

\renewcommand\thesection{\Roman{section}}

\section{Details of Annotation Cost Experiment}\label{sec:details_of_annotation}

In Section \textcolor{red}{3.1} of the main body of our paper, we report the annotation cost of four common weakly-supervised and one fully-supervised approaches. Here, we describe how we used the labeling software to generate the various kinds of annotations in Figure~\ref{Fig_SM_2}. The detailed annotation cost statistics are shown in Figure~\ref{Fig_SM_1}.

As shown in Figure~\ref{Fig_SM_2}, we use Adobe Photoshop 2019 as labeling software. Randomly-selected 100 images from the NUAA-SIRST \cite{ACM} and NUDT-SIRST \cite{DNANet} datasets are used for annotation cost evaluation. Averages annotation cost from 3 trails are reported in Figure~\ref{Fig_SM_1}. Given an input image, we first localize the small targets and then zoom in the targets located small region, as shown in Figure~\ref{Fig_SM_2} (a)-(c). Then, multiple types of label are placed in the target region by corresponding labelling manners. For single point annotation (shown in Figure~\ref{Fig_SM_2} (d)), we place the single-point label on the center of the target. For multiple points annotation (shown in Figure~\ref{Fig_SM_2} (e)), the first point is also placed on the center of the target. Then, the remaining 4 points are randomly placed in the four corners of the target. Scribble annotation is a curve that passes through the central region of the target (shown in Figure~\ref{Fig_SM_2} (f)). Bounding boxes annotation (shown in Figure~\ref{Fig_SM_2} (g)) is a box that fits tightly to the target. Pixel level annotation (shown in Figure~\ref{Fig_SM_2} (h)) is achieved by labelling every pixels in the image.

Figure~\ref{Fig_SM_1} reports the annotation cost of five kinds of annotations by re-labelling the NUAA-SIRST \cite{ACM} and  NUDT-SIRST \cite{DNANet} datasets.  We take 1.4, 3.1, 5.1, 6.5, and 11.2 seconds per target to generate single point, multiple points, scribbles, bounding boxes, and pixel-level annotations, respectively. Single-point supervision can reduce about 87\% annotation time as compared to the pixel-level annotation approach.

\begin{figure}
\centering
\includegraphics[width=0.99\linewidth]{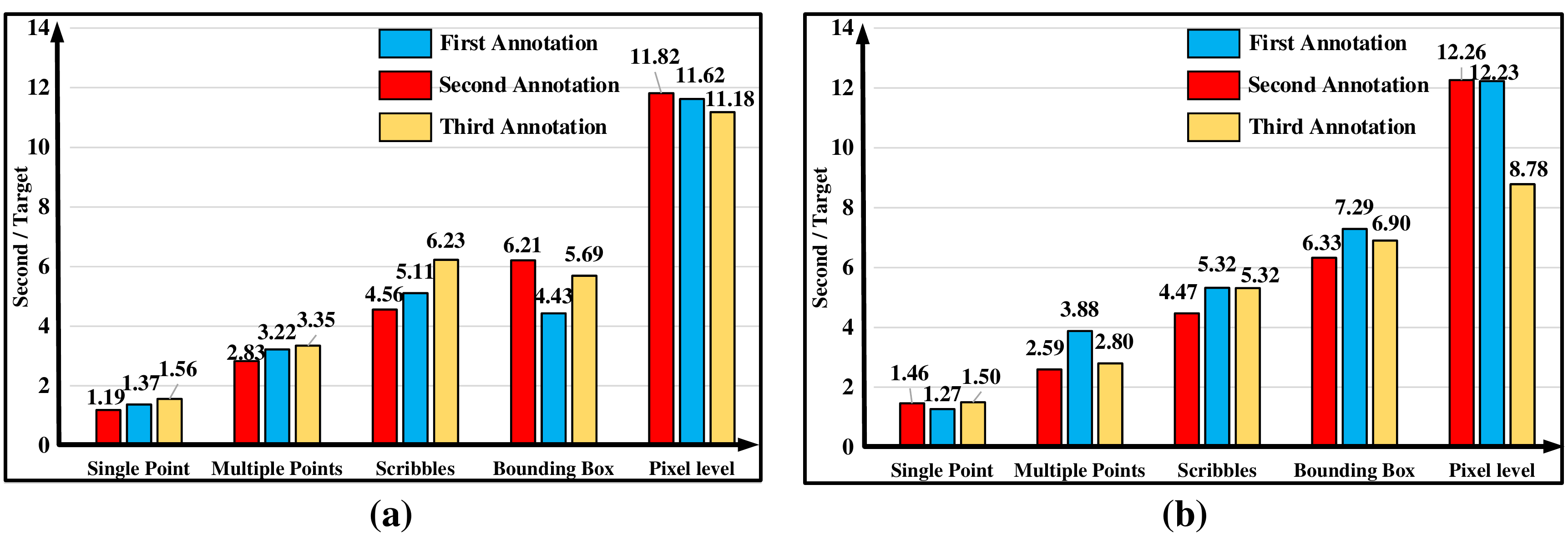}
\caption{ Annotation time cost on the (a) NUAA-SIRST and (b) NUDT-SIRST datasets. We take 1.4, 3.1, 5.1, 6.5, and 11.2 seconds per target to generate single point, multiple points, scribbles, bounding boxes, and pixel-level annotations, respectively. Averages annotation cost from 3 trails are reported.}\label{Fig_SM_1}
\end{figure}

\section{Details of Color Distance Experiment}\label{sec:details_of_Regularization}

\begin{figure*}
\centering
\renewcommand\thefigure{\Roman{figure}}
\includegraphics[width=0.83\linewidth]{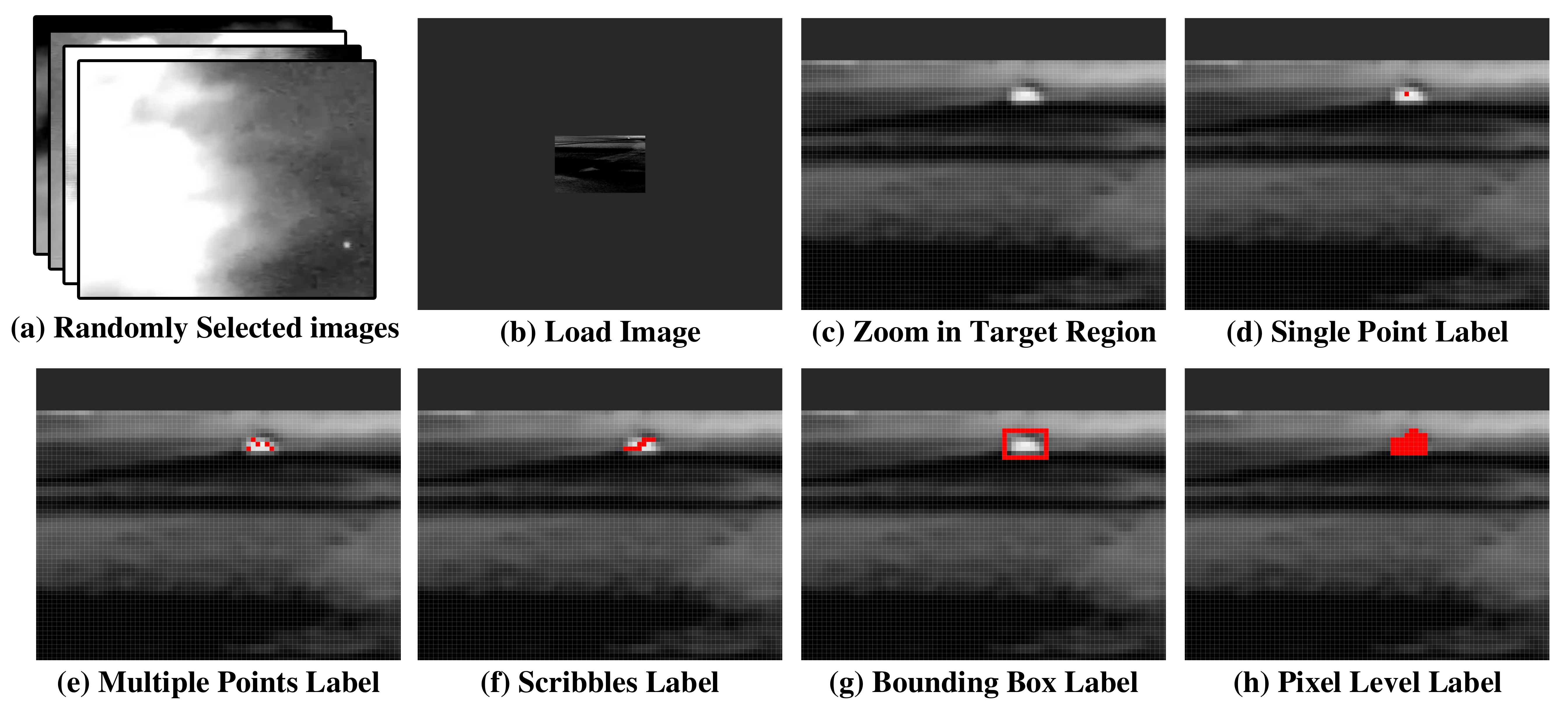}
\caption{ Detailed process of achieving four common weakly-supervised and one fully-supervised labels. Adobe Photoshop 2019 is used as labelling software.}\label{Fig_SM_2}
\vspace{-0.2cm}
\end{figure*}

In Section \textcolor{red}{3.3} of the main paper, we provide qualitative results of color distance change experiment on Figure \textcolor{red}{6} (b). Here, we provide more details for this experiment.

As described in the main body, we want to verify the opinion that random noise helps to increase the distance between background and target region. In this way, the misclustered target regions can be pushed away from the false clustering center and thus return back to the true clustering center. To support this opinion, we first calculated three kinds of color distance (i.e., maximum color distance with false clustering center, minimum color distance with false clustering center, color distance with true clustering center) and drew three kinds of curves (i.e., upper boundary of $ \Delta (\mathcal{D}_{c}(\mathcal{C}^{F}, \mathbf{\emph{M}}_{pred}))$, lower boundary of $\Delta (\mathcal{D}_{c}(\mathcal{C}^{F}, \mathbf{\emph{M}}_{pred}))$, and $\Delta (\mathcal{D}_{c}(\mathcal{C}^{T}, \mathbf{\emph{M}}_{pred}))$) with the increasing of noise intensity in Figure \textcolor{red}{6} (b).

We can observe from Figure \textcolor{red}{6} (b) that, with the increase of noise intensity, the color distance between edge areas and true clustering center (i.e., $ \Delta (\mathcal{D}_{c}(\mathcal{C}^{T}, \mathbf{\emph{M}}_{pred}))$) is always smaller than the lower boundary of $ \Delta (\mathcal{D}_{c}(\mathcal{C}^{F}, \mathbf{\emph{M}}_{pred}))$. This demonstrates that random noise with proper intensity has higher probability of pushing the target away from the false clustering center and helping them return back to the true clustering center.  Note that, the above curves are drawn by averaging the results from 10 trails. Although random noise cannot always introduce true guidance for misclustered regions in each experiment, the average result of multiple experiments presents a robust trend that random noise can effective guide the misclustered regions to return back to the true clustering center.

\section{MCLC Process on Different Datasets}\label{sec:details_of_TPM}

Here, we introduce more visual TPMs during Monte Carlo linear clustering (MCLC) process in Figure~\ref{Fig_SM_3}. Clustering results at iteration 1, 2, 20, 100 of MCLC process and the denseCRF refined pseudo masks on the NUAA-SIRST \cite{ACM}, IRSTD-1k \cite{Shape-matter}, NUDT-SIRST \cite{DNANet}, and NUDT-SIRST-sea \cite{SIRST-sea} datasets are shown in Figure~\ref{Fig_SM_3}. Although easily producing inaccurate results at the beginning of clustering (e.g., iteration number less than 20), MCLC can gradually recover a reliable clustering result.

\section{Ablation Study on Different Datasets}\label{sec:details_of_ablation}

In Section \textcolor{red}{4.3} of the main body of our paper, some ablation studies (e.g., Type and Intensity of Noise, Number of Clustering Center, Influence of Labeling Position Deviation ) were conducted on the NUAA-SIRST dataset \cite{ACM} only. Here, we present the experimental results on the other datasets (IRSTD-1k \cite{Shape-matter}, NUDT-SIRST \cite{DNANet}, and NUDT-SIRST-sea \cite{SIRST-sea}) to verify the generalization our method.

Figure~\ref{Fig_SM_4} (a1)-(d1) show the change trend of $IoU$ with respect to different noise intensity under three types of common noise (i.e., salt, pepper, and Gaussian) on four datasets. With the increase of noise intensity, the $IoU$ of MCLC with denseCRF under salt and noise increases rapidly at the beginning. After that, excessive intensity value reduces the saliency of target region and thus results in the decrease of $IoU$. Moreover, Gaussian noise causes huge performance decrease under any intensity. Similar change trend can also be found on the other datasets (IRSTD-1k, NUDT-SIRST, and NUDT-SIRST-sea). These results disclose the fact that proper type and intensity of noise are essential to MCLC on all datasets.

Figure~\ref{Fig_SM_4} (a2)-(d2) report the $IoU$ of the results generated by MCLC with denseCRF,  MCLC with fixed threshold, and LCA under different number of clustering center. It can be observed that the $IoU$ firstly shows a rapid increasing trend with the increase number of clustering center. Afterwards, the quality of TPM gradually decreases when the number of clustering center further increases. Similar change trend can be also found on the other datasets (IRSTD-1k, NUDT-SIRST, and NUDT-SIRST-sea). The above results demonstrate that inappropriate number of clustering center will result in over-small or over-large area of initial search region, and thus introduce negative effect on MCLC for all datasets.

As shown in Figures~\ref{Fig_SM_4} (a3)-(d3), with the increase of label position deviation, $IoU$ value gradually decreases, but still maintains at a high level even with five pixels deviation. Similar change trend can be also found on the other datasets (IRSTD-1k, NUDT-SIRST, and NUDT-SIRST-sea). That is because, our proposed Monte Carlo regularization method enables the model to seek for robustness representation from repetitive Monte Carlo clustering.

\section{Quantitative Results on Different Datasets}\label{sec:details_of_quantitative}

Figures~\ref{Fig_SM_5} and ~\ref{Fig_SM_6} show the qualitative results of our single-point supervised method and the compared fully-supervised methods on different SIRST datasets \cite{ACM, Shape-matter, DNANet, SIRST-sea}.

\begin{figure*}
\centering
\renewcommand\thefigure{\Roman{figure}}
\includegraphics[width=0.85\linewidth]{Fig_MCLC/More_MCLC.pdf}
\caption{Examples of target probability map and the corresponding refined pseudo masks during the MCLC process on four datasets. }\label{Fig_SM_3}
\end{figure*}

\begin{figure*}
\centering
\renewcommand\thefigure{\Roman{figure}}
\includegraphics[width=0.97\linewidth]{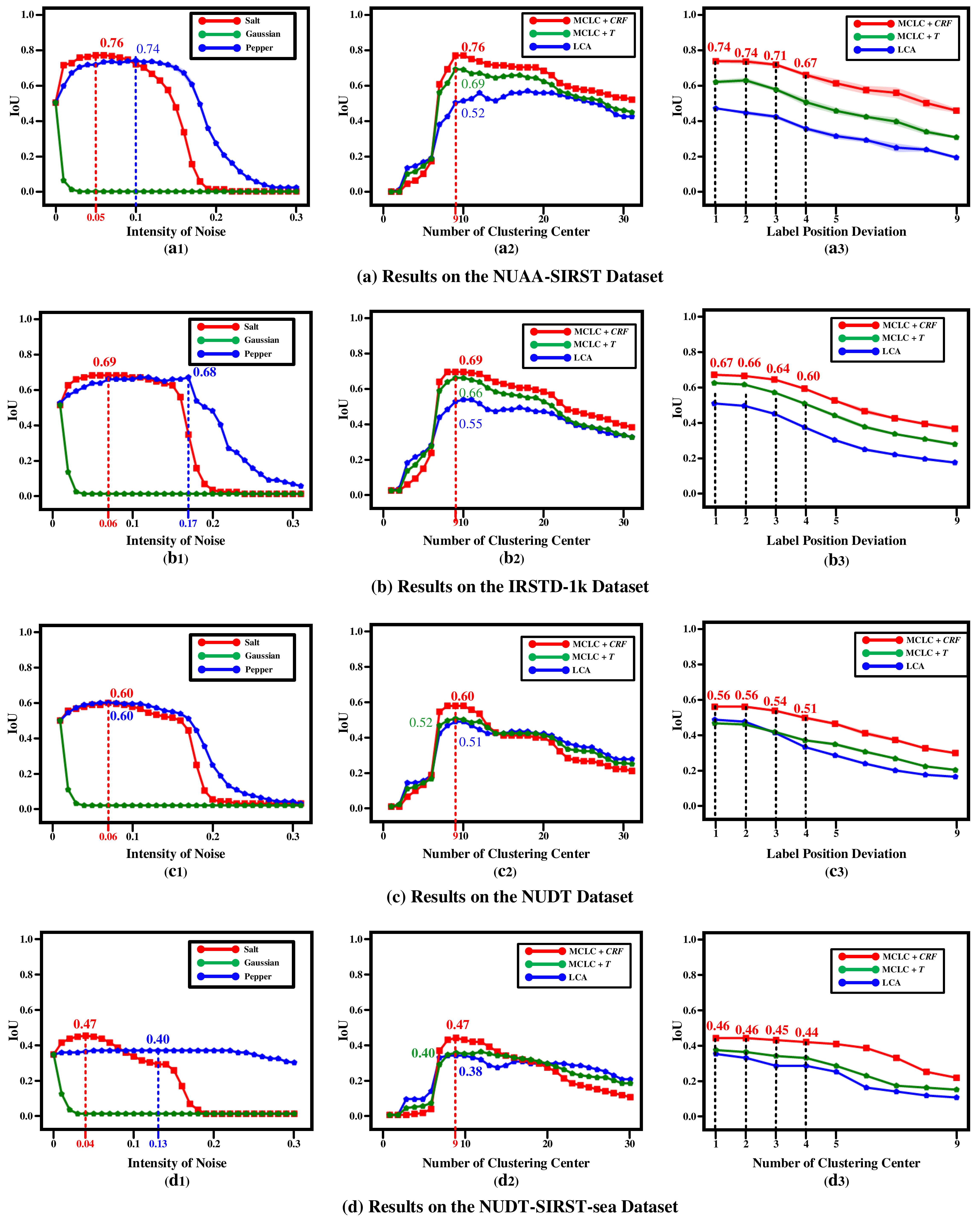}
\caption{ $IoU$ scores achieved by our method with (a1)-(d1) different types and intensity of additional noise, (a2)-(d2) different number of clustering center, and (a3)-(d3) different label position deviation on four SIRST datasets.}\label{Fig_SM_4}
\end{figure*}

\begin{figure*}
\centering
\renewcommand\thefigure{\Roman{figure}}
\includegraphics[width=0.82\linewidth]{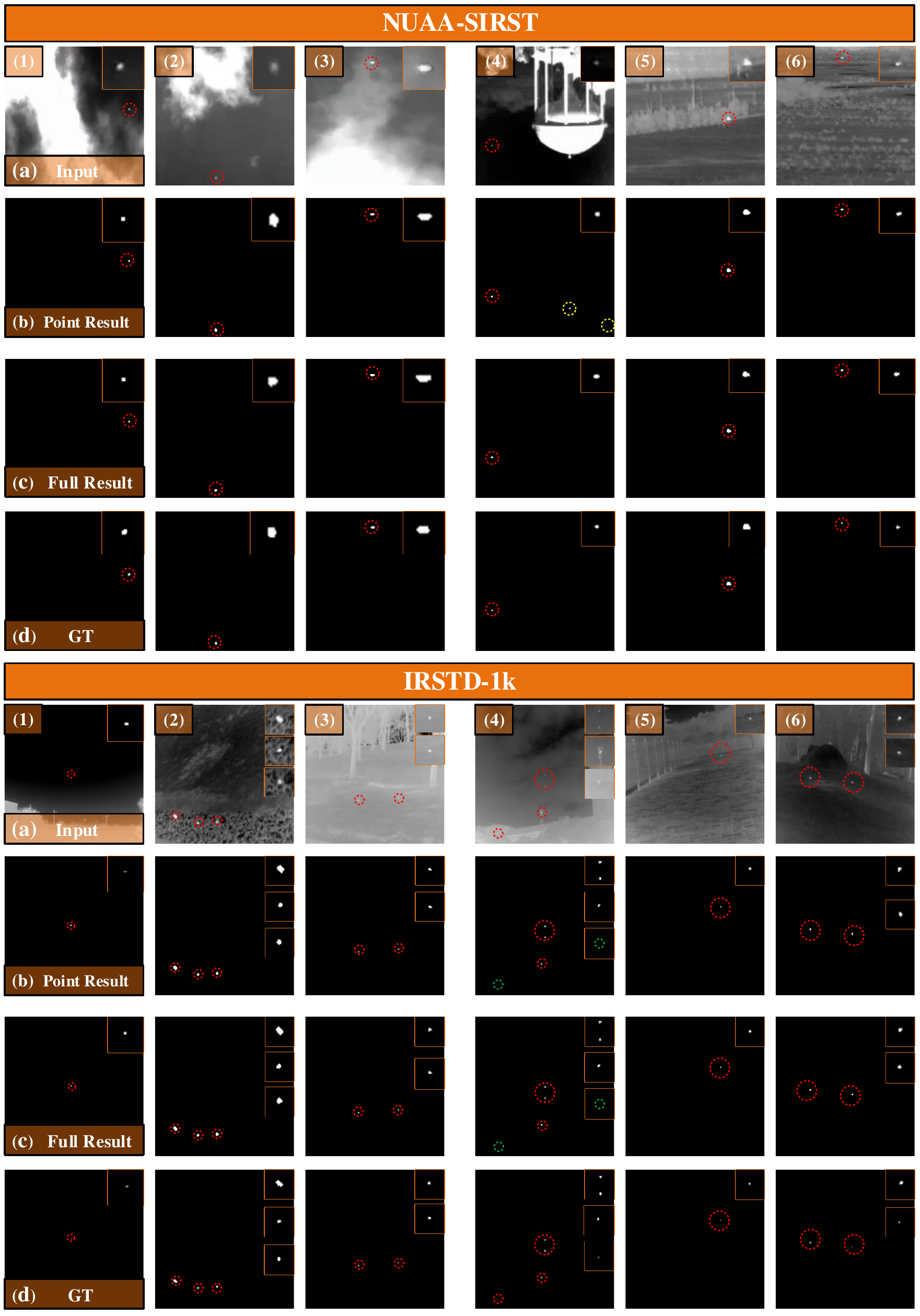}
\caption{ Qualitative results achieved by different SIRST detection methods on the NUAA-SIRST and IRSTD-1k datasets under (b) point-level supervision, (c) pixel-level supervision. For better visualization, the target area is enlarged in the right-top corner. The correctly detected target, false alarm, and miss detection areas are highlighted by red, yellow, and green dotted circles, respectively. Our MCLC enables the network to achieve comparable performance to the fully-supervised results with only single-point annotation.}\label{Fig_SM_5}
\end{figure*}

\begin{figure*}
\centering
\renewcommand\thefigure{\Roman{figure}}
\includegraphics[width=0.82\linewidth]{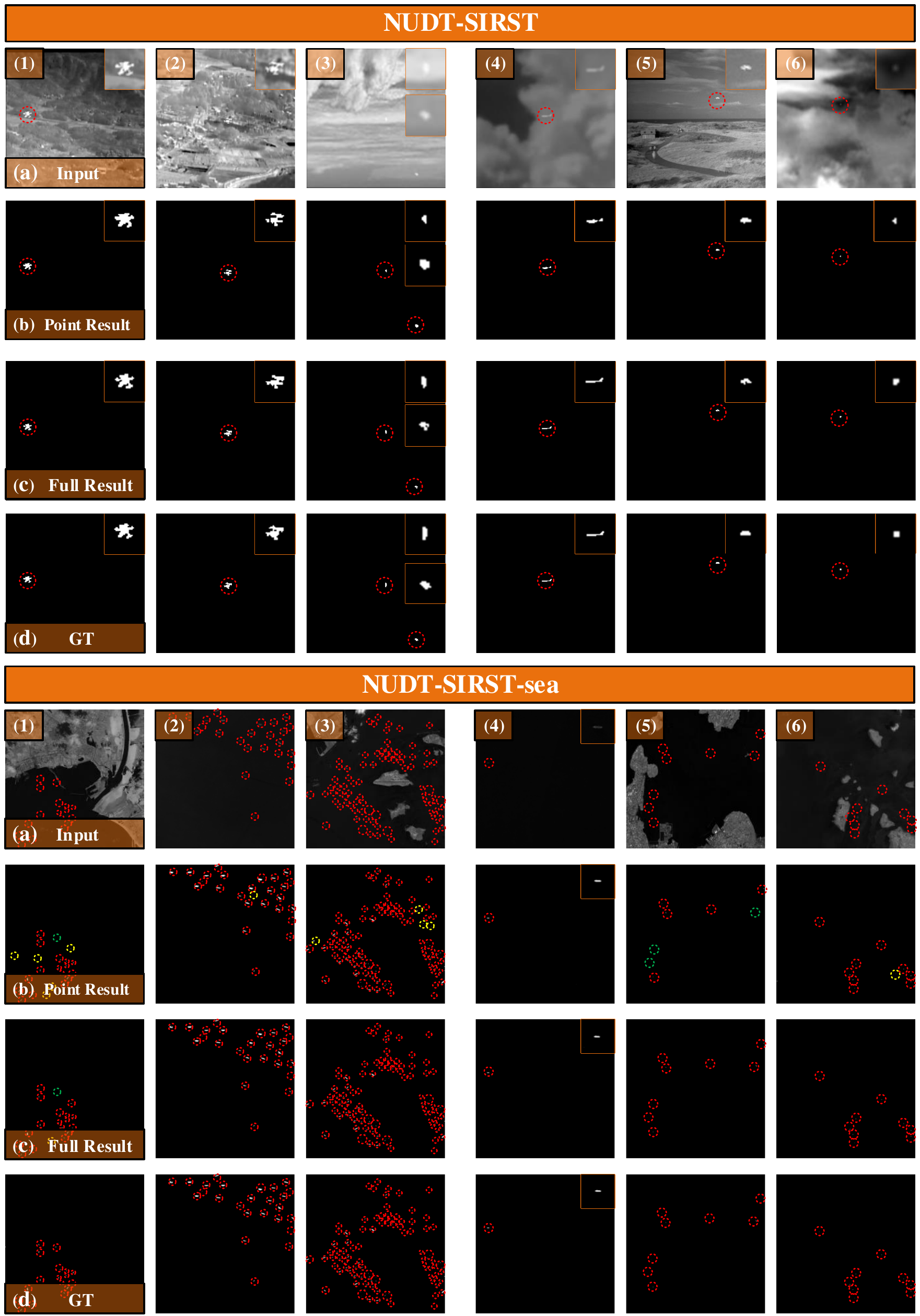}
\caption{  Qualitative results achieved by different SIRST detection methods on the NUDT-SIRST and NUDT-SIRST-sea datasets under (b) point-level supervision, (c) pixel-level supervision. For better visualization, the target area is enlarged in the right-top corner. The correctly detected target, false alarm, and miss detection areas are highlighted by red, yellow, and green dotted circles, respectively. Our MCLC enables the network to achieve comparable performance to the fully-supervised results with only single-point annotation.}\label{Fig_SM_6}
\end{figure*}
}

\end{document}